\documentclass{article}

    \PassOptionsToPackage{numbers, compress}{natbib}

 \usepackage[main, final]{neurips_2026}

\usepackage[utf8]{inputenc}
\usepackage[T1]{fontenc}
\usepackage{hyperref}
\usepackage{url}
\usepackage{booktabs}
\usepackage{amsfonts}
\usepackage{nicefrac}
\usepackage{microtype}
\usepackage{xcolor}
\usepackage{enumitem}
\usepackage{times}
\usepackage[utf8]{inputenc}

\usepackage{amsmath,amsfonts,bm}

\newtheorem{proposition}{Proposition}

\def\eqref#1{equation~\ref{#1}}

\def\1{\bm{1}}

\DeclareMathAlphabet{\mathsfit}{\encodingdefault}{\sfdefault}{m}{sl}
\SetMathAlphabet{\mathsfit}{bold}{\encodingdefault}{\sfdefault}{bx}{n}

\newcommand{\R}{\mathbb{R}}

\usepackage{hyperref}
\usepackage{url}
\usepackage{booktabs}   
\usepackage[pdftex]{graphicx}
\usepackage{microtype}
\usepackage{graphicx}
\usepackage{amsmath}
\usepackage{wrapfig}
\usepackage{float}
\usepackage{listings}
\usepackage{caption}
\usepackage{subcaption}
\definecolor{linkteal}{RGB}{0, 100, 120}

\hypersetup{
    colorlinks=true,
    linkcolor=linkteal,
    citecolor=linkteal,
    urlcolor=linkteal,
    filecolor=linkteal,
}
\title{Improved State Mixing in Higher-order and Block Diagonal Linear Recurrent Networks}

\author{%
  Igor Dubinin \\
  Natural Intelligence \\
  Ernst Strüngmann Institute \\
  \texttt{igor@nisys.ai} \\
  \And
  Antonio Orvieto\\
  ELLIS Institute Tübingen \& MPI-IS \\
  Tübingen AI Center\\
  \texttt{antonio@tue.ellis.eu}
  \And
  Felix Effenberger \\
  Natural Intelligence \\
  Ernst Strüngmann Institute \\
  \texttt{felix@nisys.ai}
}

\newtheorem{corollary}{Corollary}
\renewcommand{\arraystretch}{0.8}
\begin{document}

\maketitle

\begin{abstract}
  Linear recurrent networks (LRNNs) and linear state space models (SSMs) promise computational and memory efficiency on sequence modeling tasks, yet their diagonal state transitions limit expressivity.
Dense and/or nonlinear architectures (e.g., LSTMs) on the other hand are provably more expressive, but computationally costly.
Here, we explore how expressivity in LRNNs can be increased via richer state mixing across time and channels while maintaining competitive efficiency.
Specifically, we introduce two structured LRNN architectures: (i) Higher-order Linear Recurrent Units (H-LRU), which generalize recurrences to arbitrary order, mixing multiple past states, and (ii) Block-Diagonal LRUs (BD-LRU), which enable dense intra-block channel mixing.
To ensure stable training, we introduce a selective gate normalization scheme that allows for scalable window and block sizes.
To maintain efficiency, we implement block-diagonal parallel scans in PyTorch and Triton that keep the throughput competitive with diagonal LRNNs for moderate orders (H-LRU) and block sizes (BD-LRU). 
Consistent with prior theoretical studies on the limitations of diagonal models, we empirically demonstrate that our architectures benefit from the increased expressivity of structured state mixing, on synthetic sequence modeling and on language modeling tasks that require tracking structure.
Our results show that the structure of state mixing is a critical driver of performance in LRNNs, offering a practical pathway to closing the efficiency–expressivity gap in linear sequence models.
\end{abstract}

\section{Introduction}
\vspace{-2mm}

Recent studies have highlighted fundamental limitations of linear recurrent networks (LRNNs) by showing that the structure of the state-transition matrix results in a trade-off between efficiency and expressivity~\citep{merrill2023parallelism,cirone2024theoretical, merrill2024illusion}.
Architectures based on diagonal matrices enable an efficient implementation but are inherently limited in expressive power, while dense models are provably more expressive yet computationally prohibitive.
To bridge this gap, several LRNN architectures have been proposed: efficient structured architectures such as ones with diagonal-plus-low-rank matrices~\citep{yang2024gated,peng2025rwkv} and their products~\citep{siems2025deltaproduct}, ones based on approximations of dense matrices at test time~\citep{sun2024learning,movahedi2025fixed,von2025mesanet}, ones based on structured sparse transition matrices~\citep{terzic2026structured,terzic2026flash}, and other solutions that are \textit{de facto} equivalent to block-diagonal architectures (e.g., oscillatory blocks~\citep{rusch2024oscillatory} and complex-valued states~\citep{orvieto2023resurrecting,de2024griffin}).
Together, these studies suggest that exploring the configuration space between diagonal and dense transition matrices may yield more expressive LRNN models.

\begin{figure*}[t]
    \centering
    \includegraphics[width=\textwidth]{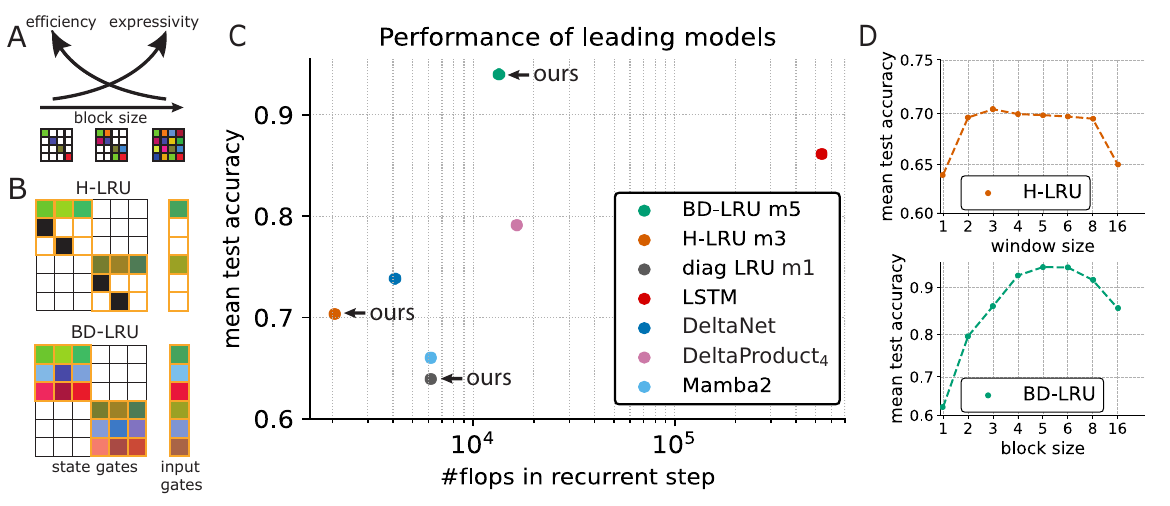}
    \caption{Structure and performance of the proposed H-LRU and BD-LRU architectures.
\textbf{A.} A schematic illustration of the theoretically predicted trade-off between expressivity and efficiency of block-diagonal linear recurrent networks.
\textbf{B.} Schematic illustration of the gating mechanisms in block-diagonal form, showing both the state gates that constitute the state-transition matrix and the input gates that act on external inputs.
The structure of the gates' selectivity is color-coded: white squares indicate fixed zero gates, black squares indicate fixed identity gates, other colors indicate active selective gates; similar color palettes indicate row-wise normalization.
\textbf{C.} Summary of the performance of the proposed and the baseline models.
The \textit{x}-axis indicates the number of FLOPs per recurrent step.
The \textit{y}-axis denotes the mean test accuracy over all considered synthetic tasks (compression, selective copying, in-context recall, permutation) of the overall best-performing model configuration (hidden size up to 6k).
Optimal hidden sizes vary between models, see also Fig.~\ref{fig:mad_scaling}.
    Note that H-LRU and BD-LRU can achieve better or matching performance than both linear and nonlinear baselines while requiring fewer FLOPs per recurrent step.
Diagonal LRU reports the best results across both H-LRU m1 and BD-LRU m1, which are identical for $m=1$.
\textbf{D.} Best performance for different window sizes $m$ (H-LRU) and block sizes $m$ (BD-LRU).}
    \label{fig:rec_flops_mad}
    \vspace{-5mm}
\end{figure*}

When designing block-diagonal recurrences, the immediate issue one faces is that of dynamical stability and forward pass normalization -- a crucial element that is well studied and discussed in diagonal LRNNs~\citep{orvieto2023resurrecting, wang2023stablessm, zucchet2024recurrent}, yet requires additional care in non-diagonal linear architectures where eigenvalues are not readily available.
Traditionally, stability has been ensured by parametrizations that keep the eigenvalues of the transition matrix on or close to the unit circle~\citep{arjovsky2016unitary,helfrich2018orthogonal}, a strategy that mitigates both vanishing and exploding gradients.
In both selective and non-selective SSMs~(designed in continuous-time),  stability is achieved by exponential parametrization, resulting from zero-order-hold discretization techniques~\citep{gu2021efficiently,gu2023mamba}.
Although several recent studies have examined block-diagonal architectures, they either focus on parametrizations of non-selective models~\citep{biegun2024rotrnn,rusch2024oscillatory}, analyze only the stability of the state-transition matrix norm~\citep{fan2023advancing}, or rely on architectures where this matrix is normalized by design~\citep{yang2024parallelizing,walker2025structured}, without fully addressing the problem of joint normalization of selective state-transition matrix and selective input gates, which has been previously shown critical for sequence modeling in diagonal LRNNs~\citep{gu2023mamba,de2024griffin}. 

Building on this line of work, we explore how to improve expressivity of LRNNs through structured selective state mixing, without compromising their computational efficiency.
We note that the goal of this paper is not to introduce a new SOTA, but to isolate and study the impact of structured state mixing on the capabilities of LRNNs.
Our contributions are as follows:
\vspace{-2mm}
\begin{itemize}[itemsep=1pt, leftmargin=*]
  \item We introduce two architectures with structured selective state mixing: (i) Higher-order Linear Recurrent Units (H-LRU) which generalize recurrences to arbitrary order, mixing multiple past states, and (ii) Block-Diagonal LRUs (BD-LRU), which enable dense intra-block channel mixing.
  \item We present empirical results on synthetic sequence modeling and language modeling tasks showing that state mixing is a critical driver of performance in LRNNs.
  \item We propose a joint selective gate normalization that ensures stable scaling with respect to block and window sizes while preserving input-dependence.
  \item We provide parallel-scan implementations in PyTorch and Triton\footnote{Code: \url{https://github.com/IgorDn8/mad-lab} and \url{https://github.com/IgorDn8/plainLM}.} that maintain high throughput for moderate block sizes, preserving the efficiency that motivates linear recurrences.
\end{itemize}

\vspace{-2mm}
\section{Higher-order and block-diagonal linear recurrent networks}

\vspace{-2mm}
\paragraph{Selective linear recurrences.}
Modern linear recurrent models~(e.g., LRU, Mamba), as well as linear attention models~(e.g., GLA, DeltaNet), exchange information between tokens by means of a recurrence
\begin{equation}
    \mathbf{h}_{t}= \mathbf{a}_{1,t} \odot \mathbf{h}_{t-1} + \mathbf{a}_{0,t} \odot \mathbf{v}_t,
    \label{eq:LRNN}
\end{equation}
where $\mathbf{h}_{t}\in\R^N$ is the hidden state computed at time $t$, $\mathbf{v}_t = \mathbf{W}_v \mathbf{x}_t$ is a linear projection of the input $\mathbf{x}_t$, and $\mathbf{a}_{1,t}, \mathbf{a}_{0,t}\in\R^N$ are the state and input gates prescribing how current information gets stored in the network state.
Throughout, $\odot$ denotes element-wise multiplication and $\times$ standard matrix multiplication.
Prior literature demonstrates that input-dependent gate selectivity is crucial for such models to achieve competitive performance~\citep{gu2023mamba}.

\vspace{-2mm}
\paragraph{Block Diagonal Recurrence.}
The most direct way to introduce state mixing into Eq.~\ref{eq:LRNN} is to replace the diagonal state-transition matrix with a block-diagonal one.
We partition the hidden state into $N$ blocks $\mathbf{h}_t^k\in\R^m$, $k\in[1,N]$, so that $\mathbf{h}_t\in\R^{Nm}$, and introduce the block-diagonal linear recurrent unit (BD-LRU), which implements dense channel mixing inside every block:
\begin{align}
\mathbf{h}_{t}^k &= \mathbf{A}^k_t \times \mathbf{h}^k_{t-1} + \mathbf{a}^k_{0,t} \odot \mathbf{v}^k_t \tag{BD-LRU} \label{eq:bdlru}
\\
\mathbf{A}^k_t &=  
\begin{bmatrix} 
a^{k}_{1,1,t} & \cdots & a^{k}_{1,m-1,t} & a^{k}_{1,m,t} \\
a^{k}_{2,1,t} & \cdots & a^{k}_{2,m-1,t} & a^{k}_{2,m,t} \\
\vdots & \ddots & \vdots & \vdots \\
a^{k}_{m,1,t} & \cdots & a^{k}_{m,m-1,t} & a^{k}_{m,m,t}\\ 
\end{bmatrix}, \nonumber
\mathbf{h}^k_{t-1}=\begin{bmatrix} h_{1,t-1}^k \\ \vdots \\ h_{m,t-1}^k \end{bmatrix}, \;
\mathbf{a}^k_{0,t} =\begin{bmatrix} a^{k}_{1,0,t} \\ \vdots \\ a^{k}_{m,0,t} \end{bmatrix} , \; \mathbf{v}^k_{t}=\begin{bmatrix} v_{1,t}^k \\ \vdots \\  v_{m,t}^k \end{bmatrix}. \nonumber
\end{align}
Here, the block size $m$ corresponds to the size of the square matrix $\mathbf{A}^k_t$, and $k$ corresponds to the block index.
All vectors $\mathbf{a}^k_{0,t},\mathbf{h}^k_{t}, \mathbf{v}^k_{t} \in \mathbb{R}^{m}$ and all matrices $\mathbf{A}^k_t \in \mathbb{R}^{m \times m}$ in BD-LRU are dense.
We refer to this mechanism as \textit{dense channel-wise} or \textit{instantaneous} state mixing.

\vspace{-2mm}
\paragraph{Higher-order Recurrence.}
As an alternative way to introduce state mixing, we generalize linear recurrences to arbitrary orders, introducing the Higher-order Linear Recurrent Units~(H-LRUs) as follows:
\begin{equation}
    \mathbf{h}_{t}= \sum_{i=1}^m \mathbf{a}_{i,t} \odot \mathbf{h}_{t-i} + \mathbf{a}_{0,t} \odot \mathbf{v}_t. \tag{H-LRU}
    \label{eq:hlru}
\end{equation}
This parametrizes the state evolution by an $m$-th order recurrence, where $m$ denotes the window size of the difference equation.
Such models are a standard tool in time series statistics for forecasting~(ARMA processes, see e.g.~\citet{hamilton2020time}) and are \textit{canonical} in systems theory, since they lead to minimal realization~(i.e., with the smallest memory size) of linear dynamical systems~\citep{glad2018control}. 
Similar structure underlies the transfer-function parametrization of state space models~\citep{parnichkun2024state}, which H-LRU extends from time-invariant to selective, input-dependent coefficients.

To see the block-diagonal structure of \ref{eq:hlru} and the connection with controllable canonical forms for transition matrices in systems theory, it is sufficient to denote by $h_{t}^k$ the $k$-th coordinate~($k\in\{1,2,\dots,N\}$) of $\mathbf{h}_{t}$ and by $a^k_{i,t}$ the $k$-th coordinate of $\mathbf{a}_{i,t}$, and to stack the $m$ most recent values of each coordinate into a vector $\mathbf{h}^k_{t-1}$.
Then H-LRU takes the same block form as BD-LRU:
\begin{equation}
\begin{aligned}
\mathbf{h}_{t}^k &= \mathbf{A}^k_t \times \mathbf{h}^k_{t-1} + \mathbf{a}^k_{0,t} \odot \mathbf{v}^k_t ,\\
\mathbf{A}^k_t &=  \begin{bmatrix} 
a^k_{1,t} & \cdots & a^{k}_{m-1,t} & a^{k}_{m,t} \\
    1 & \cdots & 0 & 0 \\
\vdots & \ddots & \vdots & \vdots \\
    0 & \cdots & 1 & 0 \\ 
\end{bmatrix},  
\mathbf{h}^k_{t-1} =\begin{bmatrix} h_{t-1}^k \\ \vdots \\ h_{t-m}^k \end{bmatrix}, \;
\mathbf{a}^k_{0,t} =\begin{bmatrix} a^{k}_{0,t} \\ \vdots \\ 0 \end{bmatrix} , \; \mathbf{v}^k_{t}=\begin{bmatrix} v_{t}^k \\ \vdots \\  0 \end{bmatrix},
\label{eq:hlru_canon}
\end{aligned}
\end{equation}
where $\mathbf{A}^k_t$ is a structured companion-like matrix which allows richer dynamic modes~(e.g. oscillatory modes)\footnote{The modes are determined by the eigenvalues of $\mathbf{A}^k_t$, i.e., the roots of its characteristic polynomial $\chi_{\mathbf{A}^k}(\lambda) = \det(\lambda I - \mathbf{A}^k)= \lambda^{m} - a^{k}_{1,t}\lambda^{m-1} - a^{k}_{2,t}\lambda^{m-2}- \cdots - a^{k}_{m-1,t}\lambda - a^{k}_{m,t}=0$.}.
In contrast to BD-LRU, only the first row of $\mathbf{A}^k_t$ contains selective gates, while the remaining rows shift past values of the same coordinate forward in time.
We refer to this mechanism as \textit{companion-structured} or \textit{temporal} state mixing.

Both architectures thus share the same block-diagonal form, with $N$ blocks of size $m$, but differ in how states are mixed inside each block: BD-LRU mixes different state components at the same time step, whereas H-LRU mixes each component with its own past values.
Per block, BD-LRU therefore has $m(m+1)$ selective gates ($m^2$ state and $m$ input gates) and H-LRU only $m+1$, so at a fixed parameter budget H-LRU affords about $m$ times more blocks.
Fig.~\ref{fig:rec_flops_mad}B illustrates the resulting selective gating mechanisms in block-diagonal form, showing both the state gates that form the state-transition matrix and the input gates applied to external inputs. 

\vspace{-2mm}
\section{Trainability via joint gate normalization}
\label{sec:norm}
\vspace{-2mm}

\begin{figure} [h]
    \includegraphics[width=\linewidth]{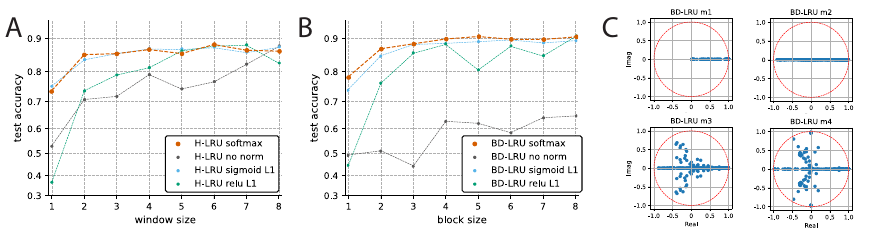}
    \centering
    \caption{Scaling of performance with window/block size on the compression task for the joint selective gate normalization with different parametrizations $f$.
Results are shown for different window/block sizes $m$ of the higher-order LRU (H-LRU) and block-diagonal LRU (BD-LRU).
\textbf{A}.
Comparison between H-LRUs.
\textbf{B}.
Comparison between BD-LRUs.
\textbf{C}.
Eigenvalue spectra of the transition matrices learned by BD-LRU on the $S5$ dataset.
BD-LRU with softmax exhibits negative eigenvalues starting from $m=2$ and complex eigenvalues from $m=3$.
Other configurations are reported in Appendix \ref{a:eigen}.}
    \label{fig:norm}
\end{figure}

While state mixing provably increases the expressivity of linear recurrences~\citep{merrill2024illusion, cirone2024theoretical}, it also makes them harder to train, which can prevent these models from reaching their theoretical capabilities.
To address this, we introduce a \textit{joint selective gate normalization}, which applies L1 normalization jointly to the state and input gates of every row of the recurrence.
We prove that it guarantees a stable recurrence and a bounded forward pass for any window or block size $m$ (Proposition~\ref{prop:norm}), and show that it greatly improves performance in practice (Fig.~\ref{fig:norm}).
The normalization does not change what the architecture can express, which is set by the structure of the transition matrix; it makes that expressivity usable by gradient descent.

\vspace{-2mm}
\paragraph{Stability alone is not enough.}
Constraining the eigenvalues of the state-transition matrix is a standard technique to prevent vanishing and exploding gradients in RNNs~\citep{pascanu2013difficulty, arjovsky2016unitary, helfrich2018orthogonal} and recent linear recurrent units~\citep{orvieto2023resurrecting} and state space models~\citep{gu2021efficiently} apply similar constraints.
However, as detailed in~\citet{orvieto2023resurrecting}, stability in a dynamical systems sense~(i.e., requiring that the eigenvalues of the hidden-to-hidden transition be less than one in absolute value) does not necessarily guarantee a properly normalized forward pass.
If the model heavily relies on past states, the accumulated hidden states can still explode unless the incoming input is scaled down proportionally.
Therefore, to maintain a bounded forward pass, both the input and forget gates must be controlled.
In purely diagonal models (Eq.~\ref{eq:LRNN}), this is typically achieved by linking the gates (as seen in S4~\citep{gu2020improving} and Mamba~\citep{gu2023mamba}) or by defining the recurrence as a direct interpolation with $\mathbf{a}_{0,t}=1-\mathbf{a}_{1,t}$, analogous to the design of GRUs~\citep{cho2014learning} and recent linear models~\citep{feng2024were}.
For block-diagonal architectures with $m>1$, no such direct coupling between a single state gate and the input gate exists, and the eigenvalues of the non-diagonal blocks $\mathbf{A}^k_t$ are not readily available in either architecture, so stability cannot be enforced by constraining them directly.

\vspace{-2mm}
\paragraph{Joint selective gate normalization.}
We generalize this normalization principle to block-diagonal and higher-order recurrences by applying L1 normalization to the raw selective gates.
For~\ref{eq:bdlru}, we apply a row-wise normalization over the hidden state gates and the input gate, while for~\ref{eq:hlru}, it is sufficient to normalize over all $m+1$ coefficients of the $m$-th order recurrence.
Denoting as $a'$s the raw gates~(linear functions of the input) before normalization, we set
\begin{equation}
\begin{aligned}
\textbf{BD-LRU:}\; a_{i,j,t} &=\frac{f(a'_{i,j,t} )}{\sum_{l=0}^m |f(a'_{i,l,t})|}; \; \; \textbf{H-LRU:}\; a_{j,t} =\frac{f(a'_{j,t})}{\sum_{l=0}^m |f(a'_{l,t})|},
\label{eq:norm}
\end{aligned}
\end{equation}
where $f(\cdot)$ is a gate parametrization function; the block index is omitted for clarity.
Note that this normalization only affects the elements inside on-diagonal blocks and has no impact on off-diagonal blocks (consisting of zero matrices).
This normalization is grounded in the following proposition, which establishes the conditions required for a stable recurrence and a normalized forward pass for any $m\ge 1$.
The proof is provided in Appendix~\ref{a:proof}.

\begin{proposition}
    \label{prop:norm}
    Consider either the H-LRU or the BD-LRU architectures, written in the block form of Eqs.~\ref{eq:bdlru} and \ref{eq:hlru_canon}.
If for any $k\in[1,N]$, the $k$-th recurrent non-diagonal block $\mathbf{h}_{t}^k = \mathbf{A}^k_t \times \mathbf{h}^k_{t-1} + \mathbf{a}^k_{0,t} \odot \mathbf{v}^k_t$ is such that the matrix $\mathcal{A}^k_t:=[\mathbf{A}^k_t, \mathbf{a}^k_{0,t}]\in\R^{m\times(m+1)}$ has the property that $\sum_{j=1}^{m+1}|(\mathcal{A}^k_t)_{i,j}| \le 1$ for every row $i\in[1,m]$, then the recurrence is stable from a dynamical systems perspective and the forward pass is normalized, meaning that $\|\mathbf{h}_{T}\|_\infty \le \max_{t\in [0,T]} \|\mathbf{v}_t\|_\infty$.
\end{proposition}

By construction, Eq.~\ref{eq:norm} satisfies the condition of Proposition~\ref{prop:norm}: in both architectures, the absolute values of the state and input gates in each row of $[\mathbf{A}^k_t, \mathbf{a}^k_{0,t}]$ sum to one.
Moreover, the normalization couples the eigenvalues of the state-transition matrix to the input gate (Corollary~\ref{cor:spec}, Appendix~\ref{a:proof}).
For the BD-LRU, the Gershgorin circle theorem applied row-wise yields $|\lambda_{i,t}| \leq \max_{i}\left( \sum_{l=1}^m |a_{i,l,t}| \right) = \max_{i} (1-|a_{i,0,t}|)$.
For the H-LRU, a Perron--Frobenius argument yields $|\lambda_{i,t}| \leq (1-|a_{0,t}|)^{1/m}$.
The normalization thus lets selective block-diagonal LRNNs balance attention to hidden states and inputs as first-order models do~\citep{orvieto2023resurrecting,de2024griffin}.
Previous studies on selective block-diagonal LRNNs only addressed the stability of the state-transition matrix~\citep{fan2023advancing}.

\vspace{-2mm}
\paragraph{Choice of parametrization.}
Beyond stability, gradient-based learning is also known to be highly sensitive to the specific choice of parametrization~\citep{zucchet2024recurrent}.
In contrast to the normalization used in non-selective block-diagonal LRNNs that rely on structured parametrizations such as discretization schemes~\citep{rusch2024oscillatory,walker2025structured}, joint parametrization of the state-transition matrices and input gate~\citep{biegun2024rotrnn}, and exponential reparametrization~\citep{orvieto2023resurrecting}, our proposed normalization is more general as it can be applied to a variety of both non-selective and selective parametrizations.
This allowed us to independently evaluate several variants of gate parametrizations that are defined by the function $f$ in Eq.~\ref{eq:norm}.
As can be seen in Fig.~\ref{fig:norm}, \textit{our joint gate normalization greatly improves performance} of both~\ref{eq:bdlru}s and~\ref{eq:hlru}s.

\vspace{-2mm}
\vspace{-2mm}
\paragraph{Gate normalization allows scaling with block size.}
Fig.~\ref{fig:norm} compares, on the compression task of Section~\ref{sec:mad}, the gate parametrizations $f$ of Eq.~\ref{eq:norm}: exponentiated gates (softmax normalization), sigmoidal gates and ReLU gates, together with unnormalized models as a baseline.
With softmax or sigmoidal normalization both architectures scale effectively with window and block size, whereas without normalization or with ReLU gates they improve much more slowly.
The largest gain occurs between a window or block size of 1 and 2; our eigenvalue analysis (Appendix~\ref{a:eigen}) attributes it to the emergence of negative eigenvalues, consistent with~\cite{grazzi2024unlocking}, and further gains from $m=3$ on to the broader range of complex eigenvalues that larger blocks admit, in line with the beneficial role of oscillatory dynamics in recurrent networks~\citep{rusch_coupled_2021, effenberger2022biology, dubinin2024fading, rusch2024oscillatory}.
For moderate block sizes ($m\in[2,5]$) softmax performs comparably to or better than sigmoidal normalization, which agrees with earlier findings that exponentiated gates benefit gradient descent~\citep{orvieto2023resurrecting,zhang2024hedgehog}, so we use it in all remaining experiments.
Appendix~\ref{a:sel} ablates non-selective parametrizations.
\vspace{-2mm}

\section{Experiments on state-tracking permutation tasks}
\vspace{-2mm}

An important property of dense recurrent networks is that one layer of such a model can easily solve inherently sequential tasks that require state-tracking.
In theory, linear diagonal networks can also solve any of these tasks, but only with a depth that grows with the sequence length.
In practice, it has been shown that they cannot effectively approximate the evolution of recurrent state with a bounded number of layers~\citep{merrill2024illusion}.
Further, it was proposed that there is a parallelism-expressivity trade-off, in which efficient parallelization comes at the expense of decreased expressivity~\citep{merrill2023parallelism}. 

To evaluate the ability of a model to perform state-tracking we use a synthetic dataset based on the symmetric group $S_n$, the group of all permutations over $n$ elements~\citep{merrill2024illusion}.
Each input is a sequence of permutations drawn from $S_n$, and the model must output at every position the composition of all permutations seen so far, which requires tracking the group state across the sequence.
We evaluated model performance on increasingly complex permutation learning tasks derived from the symmetric groups $S_2$ through $S_5$. 

\vspace{-2mm}
\paragraph{Complex permutations are challenging.}
All tested recurrent architectures (H-LRU, BD-LRU, LSTM, DeltaNet, DeltaProduct) were able to perfectly solve the $S_2$ task, which represents a uniquely simple permutation group as it is also a commutative cyclic group.
However, as the group order increases over $S_3$ to $S_5$, the non-commutative structure of the permutation tasks increasingly posed challenges for the models, see Table~\ref{tab:s_results}, Table~\ref{tab:s_results_ext}, and Table~\ref{tab:s_results_ext_eff}.
Performance of the H-LRU was found to decrease progressively with increasing group size, indicating a limited capacity for modeling compositional permutations.
Increasing the order of recurrence $m$ did not improve performance.
We conclude that a strict temporal inductive bias on the structure of the transition matrix prevents H-LRU from solving this task.
Moreover, we found that H-LRU is unable to solve our permutation tasks despite having access to negative and complex eigenvalues (see Fig.~\ref{fig:norm} and Appendix~\ref{a:eigen} for our eigenvalue analysis).
This indicates that the presence of such eigenvalues is insufficient for these tasks and highlights that the structure of state mixing plays a more critical role.

\vspace{-2mm}
\paragraph{BD-LRUs efficiently learn complex permutations.}
In contrast, BD-LRU with moderate block sizes was able to successfully solve all permutation tasks for all group sizes, matching the performance of LSTM and outperforming all other recurrent architectures tested.
Notably, BD-LRU with block size 5 solved the $S_5$ task using as few as 200K parameters, matching the parameter efficiency of the more computationally demanding, nonlinear LSTM.
Furthermore, we found that BD-LRUs are also sample-efficient in learning permutations, outperforming even LSTM in the regime of limited training data.
We note that in our low training token regime DeltaProduct$_4$ fails to learn the $S_5$ dataset.
However, when the amount of training data approaches that used by~\citet{siems2025deltaproduct}, it is capable of solving $S_5$ task, showing that low-rank matrices are less sample-efficient than BD-LRU.
Our findings align well with our predictions that dense blocks of BD-LRU are well-suited for implementing permutations between hidden states.
The consistent improvement with larger block sizes on permutation tasks of increasing complexity highlights the advantage of this inductive bias.

\begin{table}[H]
\centering
\setlength{\tabcolsep}{3.5pt}
\begin{tabular}{lrrrrrr}
\toprule
Models & $S_3$ (10k samples) & $S_3$ (250) & $S_4$  (50k) & $S_4$  (3k) & $S_5$ (100k) & Overall \\
\midrule
LSTM~\citep{hochreiter1997long} & \textbf{1.000} & 0.316 & \textbf{1.000} & 0.368 & \textbf{1.000} & 0.737 \\
Mamba2~\citep{dao2024transformers} & 0.657 & 0.281 & 0.429 & 0.119 & 0.261 & 0.349 \\
PD-SSM~\citep{terzic2026structured} & \textbf{1.000} & 0.273 & 0.569 & 0.154 & 0.347 & 0.469 \\
DeltaNet[-1,1]~\citep{yang2024parallelizing,grazzi2024unlocking} & \textbf{1.000} & 0.264 & 0.507 & 0.138 & 0.140 & 0.410 \\
DeltaProduct$_4$[-1,1]~\citep{siems2025deltaproduct} & \textbf{1.000} & 0.279 & \textbf{1.000} & 0.136 & 0.140 & 0.511 \\
H-LRU m5 & 0.622 & 0.318 & 0.448 & 0.194 & 0.377 & 0.392 \\
\midrule
BD-LRU m1 & 0.563 & 0.381 & 0.343 & 0.218 & 0.215 & 0.344 \\
BD-LRU m2 & \textbf{1.000} & 0.488 & 0.699 & 0.356 & 0.338 & 0.576 \\
BD-LRU m3 & \textbf{1.000} & \textbf{1.000} & \textbf{1.000} & 0.426 & 0.484 & 0.782 \\
BD-LRU m4 & \textbf{1.000} & \textbf{1.000} & \textbf{1.000} & \textbf{1.000} & 0.882 & 0.976 \\
BD-LRU m5 & \textbf{1.000} & \textbf{1.000} & \textbf{1.000} & \textbf{1.000} & \textbf{1.000} & \textbf{1.000} \\
\bottomrule
\end{tabular}
\vspace{2mm}
\caption{Model performance on permutation composition tasks.
We note that BD-LRU performance improves with block size $m$, demonstrating strong sample efficiency by solving the tasks even given limited training data.
We utilized all permutations belonging to the symmetric groups $S_3$, $S_4$, $S_5$ (6, 24, and 120 elements, respectively).
See Appendix~\ref{a:exp} for the data protocol and Tables \ref{tab:s_results_ext} and \ref{tab:s_results_ext_eff} for extended results with statistics.}
\vspace{-8mm}
\label{tab:s_results}
\end{table}

\section{Experiments on token manipulation tasks}
\label{sec:mad}
\vspace{-2mm}

While sequence modeling is often evaluated through large-scale training, recent work demonstrates that critical capabilities can be effectively assessed via targeted synthetic benchmarks \citep{arora2023zoology,poli2024mechanistic}.
Leveraging the established connection between lossless compression and generalization \citep{shannon1948mathematical,deletang2023language,gu2025tradeoffs}, we first evaluate temporal information integration using the auto-encoding compression task.
Furthermore, to assess the dynamic adaptation required for general sequence modeling, we employ selective copying and in-context recall \citep{poli2024mechanistic, olsson2022context,waleffe2024empirical}.

\paragraph{Performance is limited by state mixing.}
Next, we performed experiments in which we investigated the difference between scaling the window size and the hidden size.
In these experiments we found that for both H-LRUs and BD-LRUs, the scaling with hidden size could not compensate for a lack of expressivity.
In other words, window/block size was found to be the key factor for performance, see Fig.~\ref{fig:mad_scaling}.
We also found that scaling of H-LRUs and BD-LRUs results in models that are competitive with LSTMs and achieve higher performance than other linear recurrent baselines, both diagonal ones such as Mamba2 and low-rank ones such as DeltaNet and DeltaProduct, see Table~\ref{tab:mad_results}.
In line with the observed limitations of diagonal RNNs, we found that scaling the hidden size in a Mamba2 model also had limited effect on performance, see Fig.~\ref{fig:mad_scaling}.
Notably, we also found distinct scaling behaviors for the compression and our other tasks, aligning with previous results~\citep{deletang2023language}.
In the compression (auto-encoding) task, performance saturates at $m\approx3$--$4$ and degrades for $m=8$, whereas performance on autoregressive tasks keeps scaling with block size.
Therefore, the decrease in aggregate performance for larger block sizes is substantially driven by the results on the compression task.

Our experiments show a direct trade-off between parameter efficiency and peak performance, as governed by the block and window sizes for BD-LRU and H-LRU, respectively.
Models with smaller block/window sizes saturate at lower parameter counts, demonstrating high efficiency.
In contrast, models with larger block/window sizes require a larger hidden dimension to match the performance of the smaller models, but can ultimately achieve a much higher performance.
This indicates that richer state mixing increases a model's expressive power at the expense of parameter efficiency.

\begin{table}[h]
\centering
\begin{tabular}{lrrrr}
\toprule
Models & In-Context Recall & Selective Copy & Compression & Overall\\
\midrule
LSTM & \textbf{1.000} & \textbf{1.000} & 0.749 & 0.916 \\
Mamba2 & \textbf{1.000} & 0.812 & 0.721 & 0.844 \\
DeltaNet[-1,1] & \textbf{1.000} & 0.902 & 0.765 & 0.889 \\
DeltaProduct$_4$[-1,1] & \textbf{1.000} &  \textbf{1.000} & 0.735 & 0.911 \\
\midrule
BD-LRU m1 & 0.779 & 0.835 & 0.725 & 0.780 \\
BD-LRU m2 & \textbf{1.000} & 0.964 & 0.760 & 0.908 \\
BD-LRU m3 & \textbf{1.000} & 0.984 & 0.764 & 0.916 \\
BD-LRU m4 & \textbf{1.000} & 0.985 & \textbf{0.786} & \textbf{0.924} \\
BD-LRU m8 & \textbf{1.000} & 0.996 & 0.748 & 0.915 \\
\midrule
H-LRU m1 & 0.783 & 0.847 & 0.758 & 0.796 \\
H-LRU m2 & 0.998 & 0.853 & 0.772 & 0.874 \\
H-LRU m3 & \textbf{1.000} & 0.853 & 0.773 & 0.875 \\
H-LRU m4 & \textbf{1.000} & 0.845 & 0.774 & 0.873 \\
H-LRU m8 & \textbf{1.000} & 0.807 & 0.766 & 0.858 \\
\bottomrule
\end{tabular}
\vspace{2mm}
\caption{Performance on the in-context recall, selective copying and compression tasks.
The presented results are the average of best test accuracies across four configurations of the corresponding synthetic dataset with different vocabulary sizes, sequence lengths and number of training examples (best across 5 seeds).
Results are shown for different window (H-LRU) and block sizes (BD-LRU) $m$.
The overall performance of our models consistently improves with window/block size up to approximately 3–5, after which the gains saturate.
All models are single-layer configurations with a maximum overall hidden dimension of 6144.
See Table~\ref{tab:mad_results_ext} for extended results.}
\vspace{-4mm}
\label{tab:mad_results}
\end{table}

\vspace{-2mm}
\paragraph{H-LRUs are parameter efficient.}
We also found that in the compression task which does not require complex token manipulation, H-LRU demonstrated the most parameter efficient scaling with hidden size, achieving accuracies not accessible to Mamba2 and LSTM of similar sizes (in terms of the number of trainable parameters), see Fig.~\ref{fig:mad_scaling}.
This aligns well with our predictions that the inductive bias introduced by extended temporal mixing results in hidden representations with better compression capabilities.

\vspace{-2mm}
\paragraph{BD-LRUs are expressive across tasks.}
In contrast to the compression task, the selective copying task requires more extensive token manipulation.
We found that the performance of BD-LRUs scales more favorably with hidden size than that of H-LRUs.
Furthermore, BD-LRUs were able to outperform Mamba2 and DeltaNet, achieving performance that is competitive with LSTMs and DeltaProduct.
At the same time, BD-LRUs achieved the best performance in the compression task as well.
Overall, the introduced normalization scheme allows BD-LRUs to efficiently utilize the expressivity of their dense block-diagonal structure to approximate a variety of mixing patterns and to achieve the best overall results on our set of synthetic tasks.
\vspace{-2mm}
\paragraph{Further tasks.}
The remaining MAD tasks (memorization, fuzzy and noisy recall; Table~\ref{tab:mad_extra}) and the Chomsky-hierarchy tasks (Appendix~\ref{a:chom}) confirm these conclusions: BD-LRU solves parity and cycle navigation from the same block sizes as $S_3$ and $S_5$, and noisy recall is solved by both architectures at moderate block or window sizes but not by the diagonal model.

\section{Experiments on language modeling}
\label{sec:lm}
\begin{figure*}[h]
    \centering
    \includegraphics[width=\linewidth]{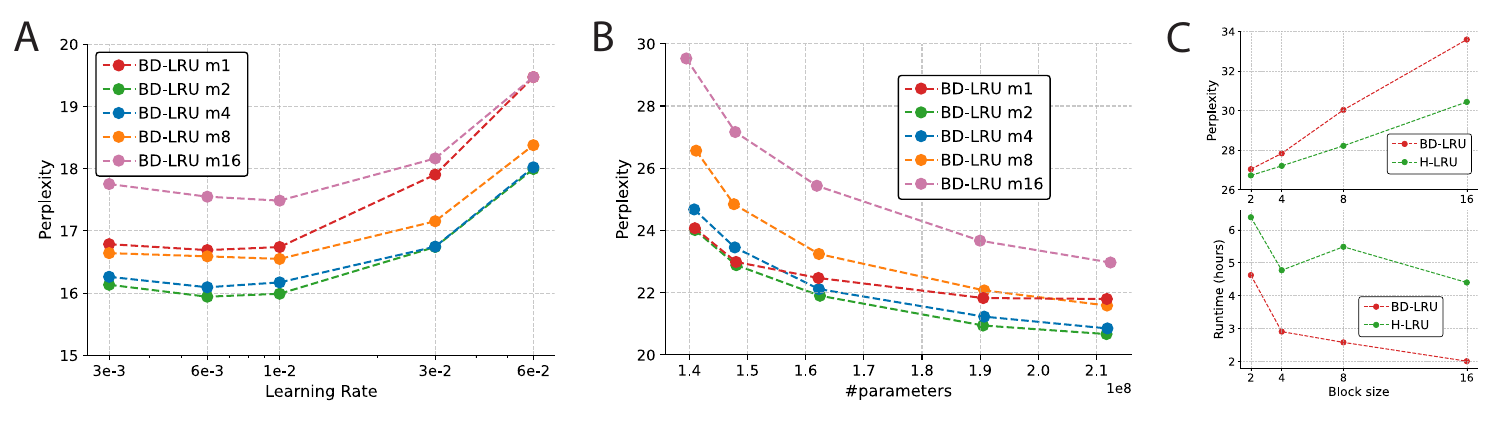}
    \caption{Language modeling on FineWeb-Edu. \textbf{A.} Best perplexity of 210M-parameter BD-LRU models (10B tokens) across learning rates. \textbf{B.} Scaling with hidden size (2.5B tokens): moderate block sizes keep improving where the diagonal model saturates. \textbf{C.} Runtime and perplexity of 140M-parameter models; matching a BD-LRU parameter budget needs an $m$ times larger H-LRU hidden size.}
    \label{fig:lm}
    \vspace{-2mm}
\end{figure*}

\vspace{-2mm}
\paragraph{Moderate block sizes scale best.}
Our language modeling experiments on FineWeb-Edu ($\sim$210M-parameter models trained on 10B tokens; setup in Appendix~\ref{a:exp}) corroborate the synthetic results, see Fig.~\ref{fig:lm}.
Among parameter-matched BD-LRU models trained on 10B tokens, moderate block sizes reach lower perplexity than the diagonal model, while the largest block size, despite being more expressive, falls behind (Fig.~\ref{fig:lm}A).
As the hidden size grows, moderate block sizes keep improving, whereas the diagonal model saturates early (Fig.~\ref{fig:lm}B).
H-LRU is more parameter-efficient than BD-LRU (Fig.~\ref{fig:lm}C), but matching the parameter budget of a BD-LRU requires a hidden dimension that is larger by the window size, which lowers throughput and raises memory by about the same factor (Sec.~\ref{sec:impl} and Appendix~\ref{a:impl}); H-LRU is therefore the more costly architecture to scale.

\vspace{-2mm}
\paragraph{Perplexity rewards retrieval rather than state tracking.}
Mamba2~\citep{dao2024transformers} and DeltaNet~\citep{yang2024parallelizing} reach lower perplexity than our architectures (Table~\ref{tab:llm_eval}).
We attribute this to mechanisms that target what perplexity on web text mostly rewards, namely local structure and the retrieval of previously seen associations, rather than state tracking: Mamba2's input-selective gating implements adaptive retention over a large state, and DeltaNet's delta rule performs targeted key-value overwriting, a fast associative memory that serves in-context recall directly.
Neither extends the expressivity class of the model, and tokens whose prediction requires maintaining a latent state are a small fraction of web corpora.
Other architectures point the same way.
Adding Householder factors to DeltaNet improves state tracking substantially while barely changing perplexity~\citep{siems2025deltaproduct}; at our scale and matched parameter count it even raises perplexity slightly, since the additional factors are paid for with a smaller state expansion.
The LSTM, the most expressive model we train, is the weakest language model, consistent with LSTMs matching Transformers on early tokens but not on later ones~\citep{kaplan2020scaling}, a deficit in the use of long context rather than in its computational power.

\begin{table}[h]
\centering
\small
\setlength{\tabcolsep}{1.5pt}
\begin{tabular}{lrrrrrrrrr}
\toprule
& \multicolumn{2}{c}{PPL $\downarrow$} & & & & & & & \\
\cmidrule(lr){2-3}
Model & FineWeb & WikiText & NIAH $\uparrow$ & Anagram $\uparrow$ & Markov $\uparrow$ & Linear $\uparrow$ & ARC-E $\uparrow$ & bAbI $\uparrow$ & PIQA $\uparrow$ \\
\midrule
BD-LRU/H-LRU m1 & 20.7 & 49.2 & 0.280 & 0.611 & 0.230 & 0.604 & 0.484 & 0.357 & 0.659 \\
\midrule
H-LRU m2  & 19.3 & 41.6 & 0.505 & 0.670 & 0.230 & 0.743 & 0.484 & \textbf{0.376} & 0.660 \\
H-LRU m4  & 19.6 & 43.8 & 0.370 & \textbf{0.698} & 0.440 & 0.743 & 0.491 & 0.338 & \textbf{0.664} \\
H-LRU m8  & 20.5 & 48.5 & 0.290 & 0.644 & 0.257 & \textbf{0.799} & 0.486 & 0.338 & 0.658 \\
\midrule
BD-LRU m2 & 19.8 & 46.0 & 0.285 & 0.639 & 0.253 & 0.688 & 0.470 & 0.331 & 0.650 \\
BD-LRU m4 & 20.4 & 47.3 & 0.275 & 0.606 & \textbf{0.513} & 0.639 & 0.495 & 0.332 & 0.658 \\
BD-LRU m8 & 20.5 & 49.9 & 0.275 & 0.632 & 0.397 & 0.424 & 0.482 & 0.349 & 0.649 \\
\midrule
LSTM      & 31.6 & 106.9 & 0.230 & 0.521 & 0.303 & 0.646 & 0.430 & 0.320 & 0.609 \\
Mamba2    & 18.7 & 40.7 & 0.285 & 0.627 & 0.490 & 0.639 & 0.512 & 0.339 & 0.663 \\
DeltaNet[-1,1]  & \textbf{16.6} & \textbf{24.3} & \textbf{0.990} & 0.332 & 0.323 & 0.514 & \textbf{0.525} & 0.319 & 0.658 \\
DeltaProduct$_4$[-1,1] & 17.2 & 27.1 & \textbf{0.990} & 0.408 & 0.330 & 0.542 & \textbf{0.525} & 0.336 & 0.661 \\
\midrule
\textit{Chance} & -- & -- & \textit{0.25} & \textit{0.50} & \textit{0.25} & \textit{0.50} & \textit{0.25} & \textit{$\sim$0.25} & \textit{0.50} \\
\bottomrule
\end{tabular}
\vspace{2mm}
\caption{Evaluation of $\sim$210M-parameter pretrained language models.
Perplexity is reported on the held-out FineWeb-Edu validation split (10M tokens, the training distribution) and on WikiText-2; all other columns report accuracy.
Best value per column in bold. BD-LRU/H-LRU m1 is the diagonal $m=1$ model (identical for both architectures); DeltaProduct$_4$ uses four Householder factors.}
\label{tab:llm_eval}
\end{table}

\paragraph{State mixing pays off beyond perplexity.}
Table~\ref{tab:llm_eval} therefore evaluates the pretrained models beyond perplexity: on associative retrieval (needle in a haystack, NIAH), on three structure probes (corrupted anagram detection, order-$k$ Markov prediction and linear vs.\ random; protocol in Appendix~\ref{a:exp}), and on standard zero-shot benchmarks.
The delta-rule models are the two best language models and nearly perfect on needle-in-a-haystack retrieval, yet they perform at or below chance on anagram detection and on linear vs.\ random.
The anagram task penalizes exact associative copying, the very operation behind strong retrieval, because the answer depends on tracking the multiset of letters rather than on reproducing a sequence.
Our models, in contrast, are well above chance on all three structure probes, and at essentially the same perplexity as Mamba2, H-LRU with a window of two nearly doubles Mamba2's retrieval accuracy.
Strength on perplexity and retrieval thus does not imply the ability to track structure, in agreement with our synthetic results; the commonsense and question-answering benchmarks do not separate the architectures at this scale.

\vspace{-2mm}
\begin{wraptable}{r}{0.44\textwidth}
\vspace{-4mm}
\centering
\footnotesize
\setlength{\tabcolsep}{4pt}
\begin{tabular}{lrrr}
\toprule
Gate parametrization & $m=1$ & $m=2$ & $m=4$ \\
\midrule
Softmax ($f=\exp$) & \textbf{46.7} & \textbf{43.3} & \textbf{42.6} \\
Sigmoid ($f=\sigma$) & 46.8 & 44.9 & 44.3 \\
Linear ($f=\mathrm{id}$) & 56.1 & 47.1 & 43.6 \\
ReLU ($f=\mathrm{relu}$) & 106.5 & 92.5 & 57.0 \\
\bottomrule
\end{tabular}
\caption{FineWeb-Edu validation perplexity ($\downarrow$) of BD-LRU versus the gate parametrization $f$ in Eq.~\ref{eq:norm} and block size $m$ (8 layers, $\sim$0.6B tokens).}
\label{tab:llm_param}
\vspace{-4mm}
\end{wraptable}

\paragraph{Gate parametrization matters for perplexity.}
Table~\ref{tab:llm_param} isolates the gate parametrization $f$ of Eq.~\ref{eq:norm} on BD-LRU language models.
As on the synthetic tasks (Fig.~\ref{fig:norm}), softmax is best at every block size with sigmoid close behind, the linear normalization catches up at larger blocks, and ReLU lags far behind.
The gap between parametrizations shrinks as the block size grows, so the choice of $f$ matters most for small blocks.

\section{Implementation}
\label{sec:impl}
\vspace{-2mm}

\begin{figure}[h]
    \centering
    \includegraphics[width=\linewidth]{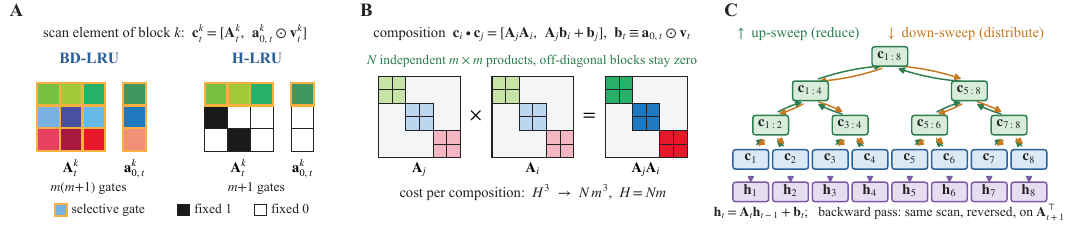}
    \caption{Structure of the block-diagonal parallel scan.
\textbf{A.} Scan element of block $k$: state gates $\mathbf{A}^k_t$ and input gates $\mathbf{a}^k_{0,t}$ (colors as in Fig.~\ref{fig:rec_flops_mad}B; $m=3$). In BD-LRU every entry is a selective gate; in H-LRU only the first row and the first input gate are selective, the other rows shift past values.
\textbf{B.} The composition multiplies block-diagonal matrices, so each step costs $N$ independent $m\times m$ products instead of one $H\times H$ product.
\textbf{C.} Blelloch up-sweep (reduce) and down-sweep (distribute) over $T$ steps, yielding all $\mathbf{h}_t$ in $\log_2 T$ parallel levels.}
    \label{fig:a_pscan_schematic}
\end{figure}

Linear recurrences can be evaluated with a parallel scan~\citep{blelloch1990prefix,sarnthein2025linearrecurrencesaccessible}, which processes a sequence of length $T$ in $\log_2 T$ parallel steps instead of $T$ sequential ones.
Fig.~\ref{fig:a_pscan_schematic} shows how we apply it to \ref{eq:hlru} and \ref{eq:bdlru}.
Each time step is represented by the pair $\mathbf{c}_t=[\mathbf{A}_t,\mathbf{b}_t]$ of its state-transition matrix and gated input $\mathbf{b}_t=\mathbf{a}_{0,t}\odot\mathbf{v}_t$, and two steps are combined by the associative composition $\mathbf{c}_i\bullet\mathbf{c}_j=[\mathbf{A}_j\mathbf{A}_i,\ \mathbf{A}_j\mathbf{b}_i+\mathbf{b}_j]$, which turns the recurrence into a prefix sum that the up- and down-sweeps of the Blelloch scan evaluate with $O(T)$ compositions (the operator is given in full in Appendix~\ref{a:impl}).
For a dense transition matrix each composition is an $H\times H$ matrix product costing $H^3$ operations, against $H$ for a diagonal matrix, so the scan is only practical for structured transitions.
The block-diagonal structure of our models keeps every product block-diagonal: a composition is $N$ independent $m\times m$ products costing $Nm^3$, which for moderate block sizes stays within an $\mathcal{O}(m^2)$ factor of a diagonal scan.
The backward pass is the same scan run in reverse on the transposed, shifted coefficients, so no second kernel is needed.
A block-diagonal parallel scan has also been used for structured linear CDEs~\citep{walker2025structured}, and our throughput scales with block size and sequence length in agreement with their findings.

\vspace{-2mm}

\vspace{-2mm}
\begin{table}[h]
\centering
\footnotesize
\begin{tabular}{lrrrrr}
\toprule
Implementation & $T=256$ & $T=1024$ & $T=4096$ & $T=16384$ & $T=65536$ \\
\midrule
Sequential (PyTorch) & 78.4 / 178 & 300 / 461 & 2383 / 1597 & -- & -- \\
Block hopscan (PyTorch) & 16.9 / 217 & 20.8 / 618 & 49.7 / 2219 & 195 / 8616 & 776 / 34199 \\
Block hopscan + \texttt{torch.compile} & 3.2 / 132 & 3.4 / 227 & 9.9 / 406 & 37.4 / 1366 & -- \\
Triton scan & 3.2 / 208 & 3.3 / 590 & 7.1 / 2109 & 26.7 / 8181 & 106 / 32456 \\
Triton scan + \texttt{torch.compile} & 3.7 / 132 & 3.8 / 228 & 4.5 / 546 & 16.5 / 1932 & 64.8 / 7478 \\
\bottomrule
\end{tabular}
\vspace{1mm}
\caption{Wall-clock time of one training step in ms and peak memory in MB (ms / MB) for five implementations of the block-diagonal scan, BD-LRU with $m=4$, $B=4$, $\sim$1.0M parameters. ``--: exceeded our budget of 5\,s per step (sequential reference, $T\ge16384$) or 165\,s of compilation (compiled PyTorch scan, $T=65536$).}
\label{tab:a_impls}
\end{table}

\vspace{-2mm}
\paragraph{Parallel scan in PyTorch and Triton.}
We first implemented this scan in PyTorch (our block hopscan, with and without \texttt{torch.compile}) and then as a Triton kernel built on \texttt{tl.associative\_scan}, which fuses the per-block compositions into a single kernel so that intermediate products never reach global memory.
Table~\ref{tab:a_impls} compares them with a sequential reference for BD-LRU at matched parameters.
The parallel scans are more than two orders of magnitude faster than the sequential reference, and our compiled Triton kernel is several times faster than our compiled PyTorch scan at moderate block sizes, while also running at the longest sequence length, where compiling the PyTorch scan exceeded our time budget.
Full sweeps over block size, batch size and sequence length for both architectures are reported in Appendix~\ref{a:impl}.

\vspace{-2mm}
\paragraph{Block size is a cheaper route to expressivity than Householder rank.}
Both our architectures and the DeltaProduct family buy expressivity through a structural parameter: the block size $m$ in our case, and the number of Householder factors $H_n$ in DeltaProduct, where $H_n=1$ is DeltaNet.
Table~\ref{tab:a_rank_cost} compares the cost of the two routes at matched parameters.
For our models it shows the predicted expressivity--efficiency trade-off: the cost of a step grows with block size, mildly for BD-LRU and strongly for H-LRU, and the trade-off becomes more pronounced as the sequence length increases (see also Fig.~\ref{fig:pscan} in the appendix).
At every matched rank BD-LRU is faster and uses less memory than the DeltaNet family, and the margin widens with sequence length, because its cost grows only weakly with $m$ whereas the DeltaNet family starts higher and grows with $H_n$.
H-LRU is ahead of the DeltaNet family up to rank four and falls behind at rank eight.
The reason is its state size rather than the scan: an H-LRU block carries $m+1$ gates instead of $m(m+1)$, so at a fixed parameter budget it affords many more blocks and a larger recurrent state, with which both the scan cost and the activation memory scale.
The ordering holds at every batch size and sequence length we tested (Appendix~\ref{a:impl}).

\vspace{-2mm}
\begin{table}[htbp]
\centering
\footnotesize
\begin{tabular}{llrrrr}
\toprule
$T$ & Model (ms / MB) & rank 1 & rank 2 & rank 4 & rank 8 \\
\midrule
4096 & BD-LRU (block size $m$) & 4.0 / 620 & 4.5 / 516 & 4.5 / 546 & 6.4 / 1447 \\
 & H-LRU (window size $m$) & 4.1 / 620 & 5.7 / 790 & 7.7 / 1333 & 17.6 / 6678 \\
 & DeltaNet / DeltaProduct ($H_n$) & 7.7 / 1203 & 8.4 / 1646 & 10.2 / 1983 & 11.8 / 2204 \\
 & LSTM & 89.2 / 299 & -- & -- & -- \\
\midrule
16384 & BD-LRU (block size $m$) & 14.4 / 2228 & 16.3 / 1821 & 16.5 / 1932 & 24.2 / 5539 \\
 & H-LRU (window size $m$) & 14.3 / 2228 & 20.7 / 2910 & 29.1 / 5083 & 69.1 / 26462 \\
 & DeltaNet / DeltaProduct ($H_n$) & 22.1 / 4512 & 33.5 / 6330 & 43.7 / 7683 & 49.2 / 8570 \\
 & LSTM & 346 / 919 & -- & -- & -- \\
\bottomrule
\end{tabular}
\vspace{1mm}
\caption{Cost of expressivity at matched parameters ($\sim$1.0M, $B=4$): step time in ms and peak memory in MB (ms / MB) versus block or window size $m$ (our models) or Householder rank $H_n$ (DeltaNet family); the LSTM has no such parameter.}
\label{tab:a_rank_cost}
\end{table}

\vspace{-2mm}
\section{Conclusion and outlook}
We introduced two linear recurrent architectures with structured selective state mixing: BD-LRU mixes the channels within a block, and H-LRU mixes a window of past states.
Our joint L1 normalization of the state and input gates keeps the forward pass stable and lets both architectures improve with block or window size.
The best structure depends on the task: BD-LRU is the best model on permutation composition and the only one that solves it with scarce data, whereas H-LRU is the most parameter-efficient on compression.
In language modeling the perplexity gains are modest, because perplexity mostly rewards retrieval, but the same models are clearly ahead on tasks that require tracking structure.
Fitting the state-mixing inductive bias to the task therefore matters more than adding width.

Our parallel scans make block-diagonal and higher-order recurrences practical: for moderate block sizes they are orders of magnitude faster than a sequential implementation and comparable to diagonal scans.
Our Triton implementation makes the block size of BD-LRU a cheaper route to expressivity than Householder factors at every matched rank.

In practice, moderate block or window sizes navigate the expressivity--efficiency trade-off, and known properties of the data should guide the choice of bias: channel mixing for state tracking, temporal mixing for compression.
Our study is limited to models of up to 210M parameters, and whether these results carry over to foundation-model scale is left for future work.
We nevertheless expect structured state mixing to be most beneficial on tasks that require tracking a latent state over many steps (e.g., entity state in long documents, program state in code, or regime switches in time series).

\section*{Acknowledgments}

We would like to thank Wolf Singer, Sajad Movahedi and Felix Sarnthein for the helpful discussions.
Antonio Orvieto acknowledges financial support from the Hector Foundation and the AI2050 Early Career Fellowship from Schmidt Sciences.

\bibliographystyle{abbrvnat}    
\bibliography{references}

\appendix
\newpage
\section{Experiments}
\label{a:exp}

\paragraph{Motivation for synthetic tasks.}

The sequence modeling capabilities of modern neural architectures are often evaluated through large-scale experiments involving models with billions of parameters and trained on trillions of tokens~\citep{kaplan2020scaling,waleffe2024empirical}.
However, recent studies have shown that many of these capabilities can  be assessed using smaller models trained on carefully designed synthetic datasets which target specific tasks that are crucial for general sequence modeling ~\citep{arora2023zoology,poli2024mechanistic}.

First, the well-established equivalence between lossless compression and probabilistic modeling suggests that models that compress well also generalize well~\citep{shannon1948mathematical,hutter2005universal}.
Indeed, recent work shows that there is a clear connection between language modeling and compression~\citep{gu2025tradeoffs}, although with some difference in scaling laws~\citep{deletang2023language}.
In light of this, we include in our evaluation a task that tests the efficiency of temporal information integration, the auto-encoding compression task from \cite{poli2024mechanistic}.

Next, general sequence modeling requires not only the ability to develop a fixed prediction algorithm, but also the capacity to adapt dynamically to changes within the input context.
Such \textit{in-context abilities} have been extensively studied and have been suggested to explain the success of the Transformer architecture \citep{olsson2022context}.
To benchmark this basic capability, we choose the selective copying and associative recall tasks that have been shown to be good indicators of the in-context abilities of sequence models \citep{arora2023zoology,poli2024mechanistic}, as well as indicators of downstream capabilities~\citep{waleffe2024empirical}. 

\paragraph{Synthetic token manipulation tasks.} 
We benchmarked our architectures using the Mechanistic Architecture Design (MAD) framework~\citep{poli2024mechanistic}, a framework for efficient model evaluation and prototyping.
The MAD protocol is motivated by the challenge of predicting how architectural choices impact performance at scale.
The working hypothesis of MAD is that an architecture's macroscopic scaling behavior can be effectively predicted by its performance on a set of microscopic, mechanistic tasks.

The benchmark consists of a diverse suite of sequence modeling challenges designed to test core token manipulation capabilities.
By evaluating models at a small, fixed computational scale, MAD produces a relative ranking of architectures that has been shown to be predictive of their compute-optimal performance in large-scale language modeling~\citep{poli2024mechanistic}.
This approach not only approximates scaling outcomes, but also provides valuable insights into the compositional skills and failure modes of a given design.

In particular, we utilize three tasks from the MAD framework:

\begin{itemize} 
\item \textbf{Compression task}.
Models are tasked to compress a random sequence of input tokens into a single aggregation token.
Then, this aggregation token is passed through an encoder MLP, the output of which is used to reconstruct the original sequence via a decoder MLP.
All models were tested using a standard encoder-decoder architecture (Embedding, Tested Model, MLP Encoder, MLP Decoder).

\item \textbf{Selective copying task}.
Models are tasked with copying tokens from one position of an input sequence to a later position of the sequence, while ignoring irrelevant noise tokens that are randomly inserted into the sequence.
This task is designed to evaluate the  ability of a model to perform selective temporal integration in the specific order of occurrence in the sequence.
All models were tested using a standard decoder-only architecture (Embedding, Tested Model, MLP Decoder).

\item \textbf{Associative recall task}.
Models are presented with an input sequence of key-value pairs and tasked with retrieving all values from the input sequence associated with the presented keys.
This task tests the ability of a model to adaptively retrieve information depending on the established in-context associations.
All models were tested using a standard decoder-only architecture (Embedding, Tested Model, MLP Decoder).
\end{itemize} 

In our experiments, each model was evaluated across four configurations: a baseline (vocabulary size: 16, sequence length: 64, training examples: 20,000) and three variations designed to probe specific failure modes.
These variations all use the same base parameters, but independently (i) increase the vocabulary size to 32, (ii) extend the sequence length to 128, or (iii) reduce the training set to 10,000 examples to test vocabulary handling, long-range capabilities, and sample efficiency, respectively.

\paragraph{Synthetic permutation tasks.}
In our experiments, we employ synthetic datasets derived from the symmetric permutation groups $S_n$, which denotes the group of all possible permutations of $n$ elements.
These groups provide a natural hierarchy of complexity: $S_2$ contains only two permutations and is fully commutative, making it relatively simple to model.
In contrast, groups with $n\geq3$ (e.g., $S_3$, $S_4$, $S_5$) are non-commutative, and their size grows factorially with $n$, which rapidly increases the difficulty of learning the underlying structure.
For instance, $S_3$, with six elements, is the smallest non-commutative group.
Geometrically, $S_3$ can be interpreted as the group of symmetries of an equilateral triangle, including both rotations and reflections.
The complexity increases substantially with $S_4$, which contains 24 elements and corresponds to the full symmetry group of a regular tetrahedron. $S_4$ introduces more intricate subgroup structures and non-trivial normal subgroups.
Extending further, $S_5$ has 120 elements and is the first symmetric group that is not solvable.
We are using all possible generating permutations for every symmetry group ($G(s_5)=120$, $G(s_4)=24$, $G(s_3)=6$).

We assess model performance on the synthetic permutation group task from \cite{merrill2024illusion}, which is designed to probe state-tracking and generalization to complex structures.
Using their toolbox, we generated datasets for the symmetric groups $S_3$, $S_4$, and $S_5$ with a fixed sequence length of 16 (for both test and train).
To evaluate sample efficiency, we created five distinct data configurations: $S_3$ (10k and 250 examples), $S_4$ (50k and 3k examples), and $S_5$ (100k examples).
The $S_5$ setting is particularly data-limited compared to the multi-million-example setups used in previous studies~\citep{siems2025deltaproduct}.
All models were tested using a standard decoder-only architecture (Embedding, Tested Model, MLP Decoder), consistent with the MAD benchmark protocol.

\paragraph{Training details on synthetic tasks.}

All models were implemented in PyTorch~\citep{paszke2019pytorch}.
For training, we follow the experimental settings of the MAD framework.
All models are trained with the AdamW optimizer~\citep{loshchilov2017decoupled} with parameters $\beta_1 = 0.9, \beta_2 = 0.999, \epsilon = 10^{-8}$ and a cosine scheduler~\citep{loshchilov2016sgdr} (minimum LR: $0.00001$), with the initial learning rate selected from $0.001$, $0.0005$, $0.0001$.
The final reported metric is the best test accuracy across all three learning rate configurations and five runs with distinct random seeds.
For training, we used NVIDIA A100 and NVIDIA H100, while  we used NVIDIA H100 for benchmarking the best throughput across models.
The estimated total compute required to perform all experiments on synthetic tasks is 16 000 GPU hours. 

\paragraph{Training details on language modeling.}
We conduct our experiments on the 10B-token sample of the FineWeb-Edu dataset~\citep{penedo2024fineweb} using the PlainLM training setup \citep{ajroldi2024plainlm}. 
All models are trained with the AdamW optimizer~\citep{loshchilov2017decoupled} with parameters $\beta_1 = 0.9, \beta_2 = 0.95, \epsilon = 10^{-8}$ and a cosine scheduler~\citep{loshchilov2016sgdr}.
All models are trained on a single NVIDIA H100 GPU, or a total of approximately 2,000 GPU hours.

\paragraph{Language modeling evaluation.}
We evaluate the $\sim$210M-parameter pretrained models of Sec.~\ref{sec:lm}, together with the baselines, at their final checkpoint and with the shared GPT-NeoX tokenizer used for pretraining.
All multiple-choice tasks are scored zero-shot in the same way: the prompt ends in a cue such as ``Answer:'' or ``Solution:'', each candidate answer is scored by the mean per-token log-probability of its continuation given the prompt (length-normalized), and the prediction is the candidate with the highest score.
Chance levels are reported in Table~\ref{tab:llm_eval}.
\begin{itemize}
    \item \textbf{Perplexity.}
We report perplexity on the held-out validation split of the FineWeb-Edu pretraining data (4,880 sequences of 2,048 tokens, about 10M tokens), which measures in-distribution language modeling quality with the exact training objective, and on the test split of WikiText-2~\citep{merity2016pointer}, a standard out-of-distribution benchmark of Wikipedia articles.
WikiText-2 is tokenized with the training tokenizer and scored with a sliding window of 2,048 tokens and a stride of 1,024 tokens.
    \item \textbf{Commonsense and question answering.}
ARC-Easy~\citep{clark2018think} contains grade-school science questions with four textual answer options (validation split, chance 0.25).
PIQA~\citep{bisk2020piqa} tests physical commonsense: given a goal, the model must choose the more sensible of two solutions (validation split, chance 0.5).
bAbI~\citep{weston2015towards} consists of synthetic reading-comprehension stories about entities, relations and multi-hop facts, followed by a question; the gold answer is scored against up to three distractor answers from the same task (validation split, chance about 0.25).
    \item \textbf{Associative retrieval.}
Needle in a haystack (NIAH)~\citep{kamradt2023needle} inserts a key--value pair into filler text and asks for the value at the end of the context, posed as a four-way multiple-choice question (chance 0.25).
We sweep context lengths from 256 to 1,920 tokens and needle positions from the beginning to the end of the context, with eight trials per cell, and report the overall accuracy.
    \item \textbf{Corrupted anagram detection.}
Given a reference letter sequence and two candidates, a true anagram and a near-anagram with $k$ letters corrupted, the model must identify the anagram (sequence lengths 4, 6 and 8, $k\in\{1,2,3,4\}$, 48 examples per cell, chance 0.5).
This requires deciding whether two strings contain the same elements in a different order, as in typo and spelling-variant detection, and is a permutation problem in the same sense as our synthetic state-tracking tasks.
    \item \textbf{Order-$k$ Markov.}
A deterministic order-$k$ chain over a small alphabet is sampled ($k\in\{1,2,3\}$, sequence length 32, 100 examples per order), and the model must predict the symbol required by the rule after observing a prefix (four-way, chance 0.25).
This probes tracking of a short symbolic history, the operation needed to follow a state machine, a protocol, or a code path.
    \item \textbf{Linear vs.\ random.}
Two unlabeled lists of integer points are shown, one lying exactly on a line $y=ax+b$ and one with the same $x$ values but shuffled $y$ values, and the model must identify the linear one (4, 6 or 8 points, 48 examples per setting, chance 0.5).
This probes detection of an exact relation in the data.
\end{itemize}
Difficulty scales with sequence length in all three structure probes, and additionally with the corruption count $k$ in the anagram task.

\section{Computational complexity}

This section provides a breakdown of the Floating Point Operations (FLOPs) required for hidden-to-hidden state transition in the recurrent architectures discussed.
For this breakdown, we define the number of blocks for BD-LRU and H-LRU as $N$ and $H$ denotes overall hidden dimension, $m \cdot N=H$ as previously.
The sequence length is denoted as $T$.
For Mamba2, the state expansion factor is denoted by $S$.
In DeltaNet and DeltaProduct$_4$, $N_h$ denotes the number of heads, $C$ denotes the number of chunks in the DeltaNet implementation, $H_n$ denotes the number of Householder transformations, and $r=1$ denotes low rank.
The calculations focus on the recurrence mechanism, omitting additional components like the input projections or gating, as they can be precomputed in advance.
A multiply-add operation is counted as 2 FLOPs.

\begin{table}[h!]
\centering
\caption{Summary of computational costs for hidden state updates.}
\label{tab:flops_summary}
\begin{tabular}{llcc}
\toprule
\textbf{Architecture} & \textbf{FLOPs per recurrent step} & \textbf{Implementation complexity} \\
\midrule
LSTM & $8H^2+25H$ & $O(TH^2)$ \\
H-LRU & $2Nm +2N$ & $O(Nm^3log(T))$ \\
BD-LRU & $2Nm^2 +2N$ & $O(Nm^3log(T))$ \\
Mamba2 & $2NS$ & $O(T(N^2+NS))$ \\
DeltaNet & $N_h(4Nr+4N)$ & $O(TCN + TN^2)$ \\
DeltaProduct$_4$ & $H_nN_h(4Nr+4N)$ & $O(H_n(TCN + TN^2))$ \\
\bottomrule
\end{tabular}
\end{table}

\newpage
\section{Implementation details and efficiency benchmarks}
\label{a:impl}
\renewcommand{\topfraction}{0.95}\renewcommand{\textfraction}{0.04}

\paragraph{Benchmark setup.}
All measurements in this section, as well as Tables~\ref{tab:a_impls} and \ref{tab:a_rank_cost} in the main text, use single-layer models with input dimension 128 and a matched parameter budget of $\sim$1.0M parameters, in bf16 on a single NVIDIA H100 80GB (PyTorch 2.13, Triton 3.7).
We sweep the batch size $B\in\{1,4,32,128\}$ and the sequence length $T\in\{256,1024,4096,16384,65536\}$, and report the median wall-clock time of one training step (forward and backward pass) over 10 iterations and 3 repeats after 3 warm-up steps, together with the peak GPU memory.
DeltaNet and DeltaProduct use the kernels of the \texttt{flash-linear-attention} library and the LSTM uses cuDNN.
For PD-SSM we use our own PyTorch associative scan, since no fused kernel is publicly available; with it, PD-SSM exceeds the GPU memory already at moderate sequence lengths, so we report it only where it fits and omit it from the comparison in the main text.
Configurations that exceed the GPU memory are marked OOM.

\paragraph{The scan in equations.}
Following the classic approach~\citep{blelloch1990prefix,sarnthein2025linearrecurrencesaccessible}, we consider a recurrence of the form
\begin{equation}
\mathbf{h}_{i+1} = \begin{cases} \mathbf{b}_{0}, & \mbox{if } \quad i = 0 \\  ( \mathbf{h}_{i} \bigotimes_v \mathbf{A}_{i}  ) \bigoplus \mathbf{b}_{i}, & \mbox{if } 0 \leq i < T  \; \end{cases},
\label{eq:rec_hopscan}
\end{equation}
where $\mathbf{h}_i,\mathbf{b}_i \in \mathbb{R}^H, \mathbf{A}_{i} \in \mathbb{R}^{H\times H}$ and  associative operators: $\bigotimes_v$ is matrix-vector multiplication, $\bigotimes_M$ is matrix-matrix multiplication and $\bigoplus$ point-wise vector summation. 

Defining the following associative operator $\bullet$ and making substitution to sequence of pairs,
\begin{equation}
\textit{HOP} =\begin{cases} \mathbf{c}_{i} = [\mathbf{A}_{i}, \mathbf{b}_{i}] \\  \mathbf{c}_{i} \bullet \mathbf{c}_{j} \equiv [\mathbf{c}_{i,A} \bigotimes_M \mathbf{c}_{j,A},(\mathbf{c}_{i,b} \bigotimes_v \mathbf{c}_{j,A})\bigoplus \mathbf{c}_{j,b}] \;
\end{cases},
\label{eq:hopscan}
\end{equation}
reduces recurrence~\ref{eq:rec_hopscan} to a classic prefix sum and allows application of up and down sweeps of the Blelloch scan.

\begin{figure}[tp]
    \centering
    \includegraphics[width=\linewidth]{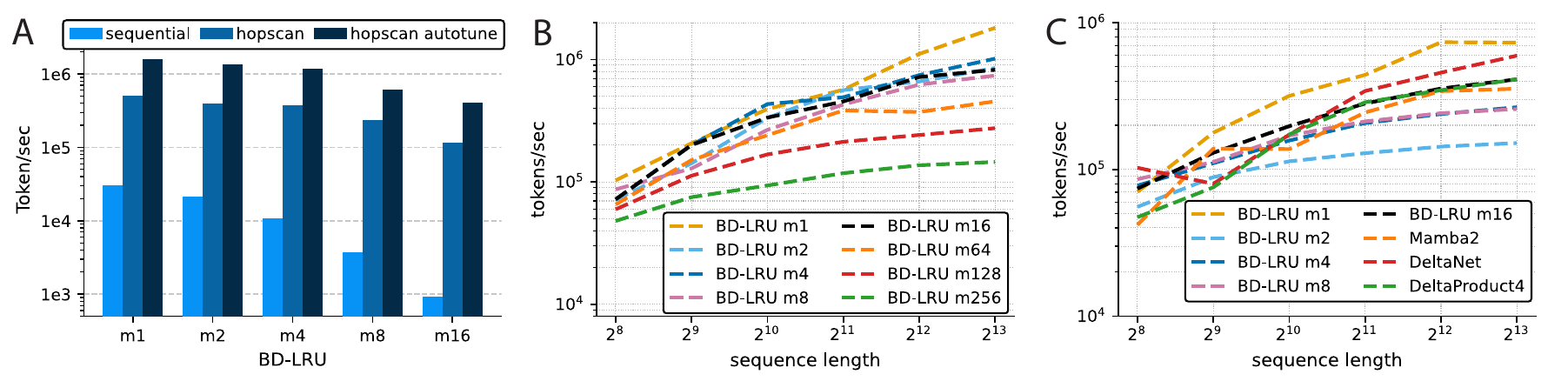}
    \caption{Training throughput of our compiled PyTorch scan (implementation (iii)) on the selective copying task.
(\textbf{A}) Sequential, parallel and compiled parallel implementations for BD-LRU with 128 blocks at sequence length 2048; H-LRU uses the same scan.
(\textbf{B}) Layers with a fixed hidden size of 768 and the number of blocks adjusted accordingly: the expressivity--efficiency trade-off grows with sequence length.
(\textbf{C}) Parameter-matched layers ($\sim$33M parameters) against LRNN baselines, with the number of blocks adjusted to keep the model size constant.}
    \label{fig:pscan}
\end{figure}

\paragraph{Scan implementations.}
We developed five implementations of the block-diagonal recurrence shared by \ref{eq:hlru} and \ref{eq:bdlru}:
(i) a sequential PyTorch reference;
(ii) our block-diagonal Blelloch scan of Eq.~\ref{eq:hopscan} in PyTorch (block hopscan);
(iii) the same scan under \texttt{torch.compile} with maximal autotuning;
(iv) a Triton kernel that implements the block-diagonal scan on top of \texttt{tl.associative\_scan} with the same composition operator, fusing the per-block compositions into a single kernel so that intermediate products are not materialized in global memory; and
(v) the Triton kernel under \texttt{torch.compile}.
Implementation (iii) produced the throughput results of Fig.~\ref{fig:pscan}; Tables~\ref{tab:a_impls} and \ref{tab:a_rank_cost} and all other figures in this appendix use implementation (v).
Fig.~\ref{fig:a_code} shows the core of implementations (ii) and (iv): the PyTorch block hopscan with its composition operator and the backward pass as a reverse scan, and the Triton kernel for $m=2$, in which the entries of $[\mathbf{A}_t,\mathbf{b}_t]$ ride along as separate inputs of \texttt{tl.associative\_scan} and the combine function is the explicit block product.
The full code, including the $m=4$ kernel, the dispatcher and the fused backward pass, is available in our repository.

\begin{figure}[tp]
    \centering
    \includegraphics[width=\linewidth]{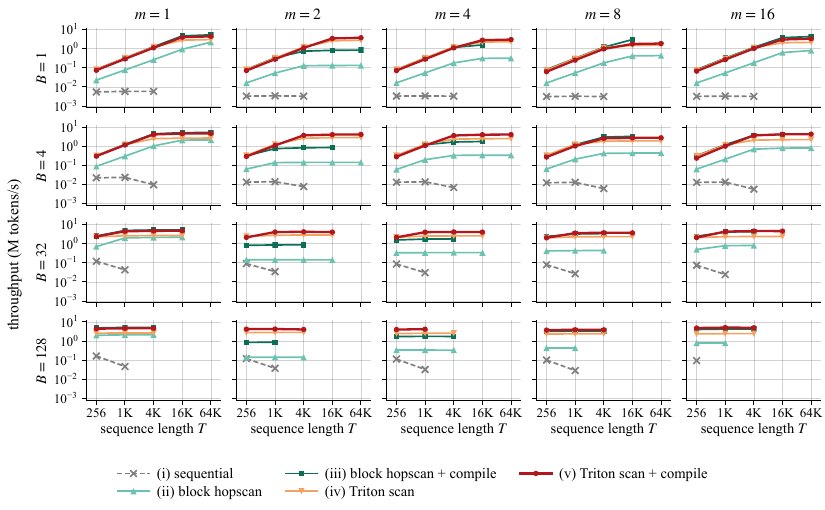}
    \caption{Training throughput of the five implementations (i)--(v) of the block-diagonal scan for BD-LRU as a function of sequence length $T$, for block sizes $m\in\{1,2,4,8,16\}$ (columns) and batch sizes $B\in\{1,4,32,128\}$ (rows), at $\sim$1.0M parameters (median over repeats). Missing points did not fit into GPU memory or exceeded our time budget (5\,s per training step or 165\,s of compilation).}
    \label{fig:a_scan_bdlru}
\end{figure}

\begin{figure}[tp]
    \centering
    \includegraphics[width=\linewidth]{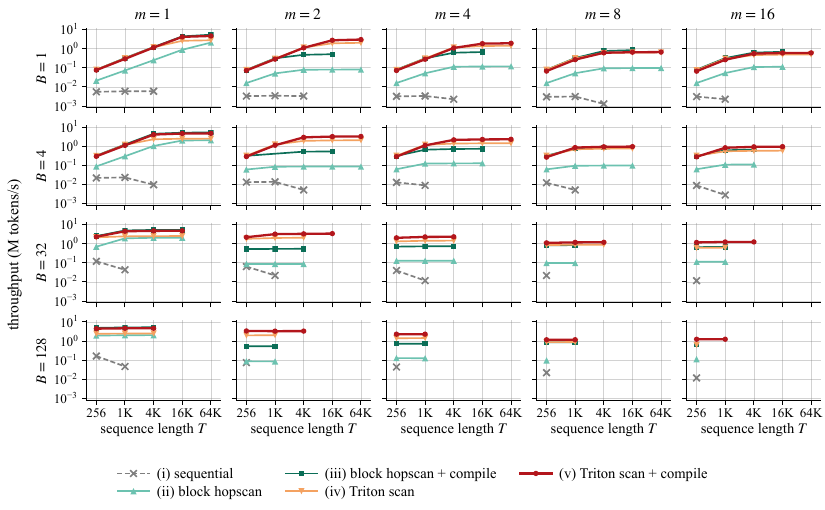}
    \caption{Training throughput of the five implementations of the block-diagonal scan for H-LRU, in the same layout as Fig.~\ref{fig:a_scan_bdlru}.}
    \label{fig:a_scan_hlru}
\end{figure}

Fig.~\ref{fig:a_scan_bdlru} and Fig.~\ref{fig:a_scan_hlru} extend Table~\ref{tab:a_impls} to all block sizes, batch sizes and sequence lengths for BD-LRU and H-LRU, respectively.
The parallel scans improve on the sequential reference by more than two orders of magnitude (2383 ms against 4.5 ms per step for BD-LRU with $m=4$ at $T=4096$).
Relative to our compiled block hopscan, the compiled Triton scan is 2.2$\times$ faster at $m=4$ and 4.5$\times$ faster at $m=2$ for BD-LRU (3.0$\times$ and 5.6$\times$ for H-LRU), with the margin growing slightly with sequence length and batch size; compiling the PyTorch scan exceeded our time budget at $T=65536$, where the Triton kernel runs at 64.8 ms per step.
For $m=1$ the two paths are equivalent, since the diagonal scan is already handled optimally by PyTorch.
For the largest block sizes the picture is architecture dependent: for BD-LRU at $m=8$ and $m=16$ the compiled PyTorch scan is on par with the Triton kernel (within 15\%), whereas for H-LRU the Triton kernel retains a $1.1$--$1.4\times$ advantage.
Peak memory of the compiled Triton scan is comparable to that of the compiled PyTorch scan and about $4\times$ below the uncompiled scans (546 MB against 2219 MB at $T=4096$).

Fig.~\ref{fig:pscan} reports the throughput of the compiled PyTorch scan in three further settings.
With 128 blocks at sequence length 2048, where runtime reflects algorithmic complexity more than GPU characteristics, throughput decreases with block size as predicted (Fig.~\ref{fig:pscan}A); at a fixed hidden size of 768 the trade-off becomes more pronounced as the sequence length grows, and block sizes above 16 lose throughput quickly at long sequences (Fig.~\ref{fig:pscan}B).
For parameter-matched layers of $\sim$33M parameters BD-LRU is competitive with the LRNN baselines (Fig.~\ref{fig:pscan}C); at this scale the differences between block sizes mostly reflect how well a given block size maps onto the GPU under compilation with maximal autotuning, and $m=16$ is in fact the most efficient despite its higher theoretical cost.

\begin{figure}[tp]
    \centering
    \includegraphics[width=\linewidth]{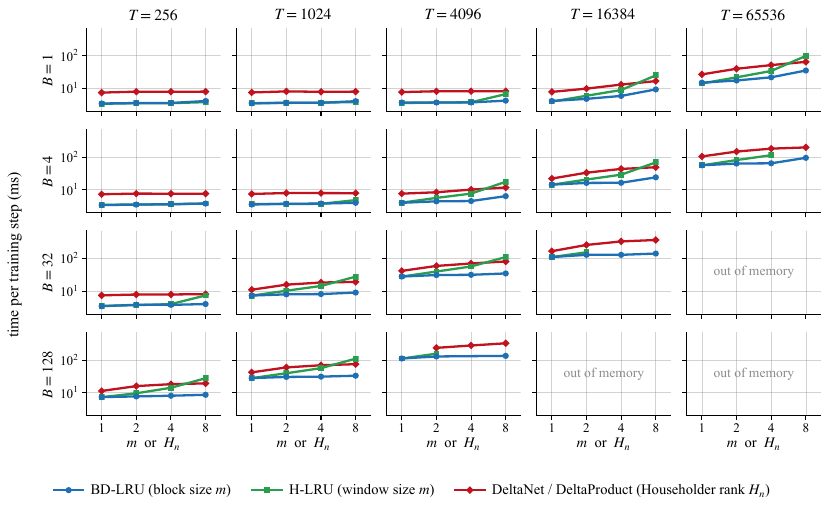}
    \caption{Wall-clock time per training step as a function of the expressivity parameter (block size $m$ for BD-LRU, window size $m$ for H-LRU, Householder rank $H_n$ for DeltaNet/DeltaProduct), at matched parameters ($\sim$1.0M) for sequence lengths $T\in\{256,\dots,65536\}$ (columns) and batch sizes $B\in\{1,4,32,128\}$ (rows), using implementation (v) for BD-LRU and H-LRU. Missing points and empty panels did not fit into GPU memory (DeltaNet at $B=128$, $T=4096$ failed with a kernel error).}
    \label{fig:a_rank_latency}
\end{figure}

\begin{figure}[tp]
    \centering
    \includegraphics[width=\linewidth]{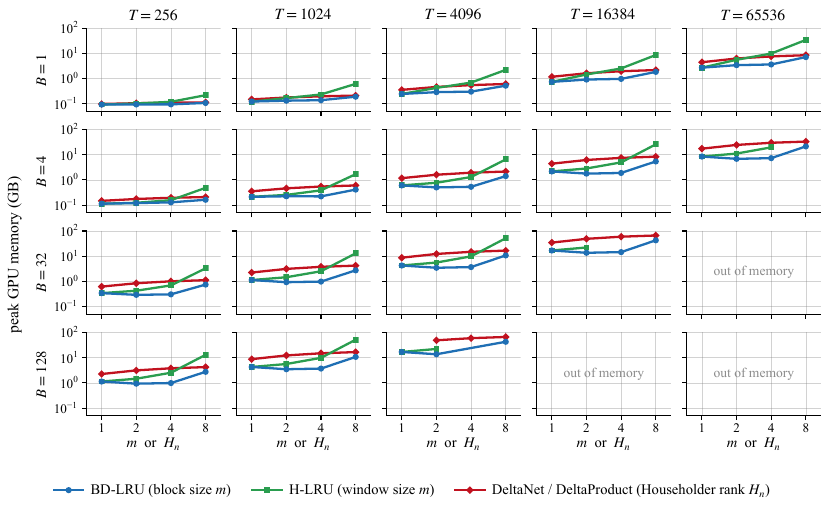}
    \caption{Peak GPU memory per training step in the same layout as Fig.~\ref{fig:a_rank_latency}.}
    \label{fig:a_rank_memory}
\end{figure}

\paragraph{Block size versus Householder rank.}
Fig.~\ref{fig:a_rank_latency} and Fig.~\ref{fig:a_rank_memory} extend Table~\ref{tab:a_rank_cost} to all batch sizes and sequence lengths.
At every matched rank BD-LRU is both faster and leaner than the corresponding member of the DeltaNet family, at both sequence lengths of Table~\ref{tab:a_rank_cost}: at rank 4 it is 2.2$\times$ faster with 3.6$\times$ less memory at $T=4096$ and 2.7$\times$ faster with 4.0$\times$ less memory at $T=16384$, and at rank 8 it is 1.8--2.0$\times$ faster with 1.5$\times$ less memory.
H-LRU is ahead of the DeltaNet family up to rank 4 and falls behind at rank 8.
The ordering holds across all batch sizes and sequence lengths we tested, and the margin widens with $T$: the cost of BD-LRU grows only weakly with $m$, whereas the DeltaNet family starts from a higher cost at rank 1 and grows with $H_n$.
PD-SSM, whose sparse transition structure makes memory grow quickly with sequence length in the absence of a fused kernel, requires 97 ms and 22.5 GB per step already at $T=1024$ and exceeds the GPU memory at $T\ge4096$ for $B=4$, which is why it does not appear in Table~\ref{tab:a_rank_cost}.

\paragraph{H-LRU versus BD-LRU at matched parameters.}
At matched parameter count H-LRU is slower and uses more memory than BD-LRU for every $m>1$: at $T=4096$ it is 1.3$\times$, 1.7$\times$ and 2.8$\times$ slower at $m=2$, $4$ and $8$, respectively (Table~\ref{tab:a_rank_cost}).
A BD-LRU block carries $m(m+1)$ selective gates, whereas an H-LRU block carries only $m+1$, so for a fixed parameter budget H-LRU affords many more blocks ($N=778$ against $N=277$ at $m=4$) and its augmented state of size $N\cdot m$ is correspondingly larger: 1.7$\times$, 2.8$\times$ and 4.9$\times$ larger than that of BD-LRU at $m=2$, $4$ and $8$.
Both the scan cost and the activation memory scale with this state, which is why H-LRU runs out of memory at $T=65536$ for $m\ge8$ (Fig.~\ref{fig:a_rank_latency}), and why H-LRU is more expensive to scale in language modeling than BD-LRU despite its parameter efficiency (Sec.~\ref{sec:lm} and Fig.~\ref{fig:lm}C).

\begin{figure}[p]
\noindent\textbf{(ii) Block hopscan in PyTorch.}
\begin{lstlisting}
import torch
from torch._higher_order_ops.associative_scan import associative_scan

# inputs: gated inputs b_t, shape (B, T, N, m); coeffs: block transitions A_t, shape (B, T, N, m, m)
def scan_hop_fwd(inputs, coeffs, reverse=False):
    def op(acc, curr):  # composition c_i . c_j = [A_j A_i, A_j b_i + b_j]
        c = torch.einsum('bcij,bcjk->bcik', curr['c'], acc['c'])
        x = curr['x'] + torch.einsum('bcij,bcj->bci', curr['c'], acc['x'])
        return dict(x=x, c=c)
    return associative_scan(op, dict(x=inputs, c=coeffs), dim=1,
                            reverse=reverse, combine_mode='generic')['x']

# backward pass: the same scan run in reverse on the shifted, transposed coefficients
# (shift(x, k) rolls x along time by k steps without wrap-around, filling with zeros)
def scan_hop_bwd(d_outputs, coeffs, outputs, reverse=False):
    coeffs_bwd = shift(coeffs, -1 if not reverse else 1).permute(0, 1, 2, 4, 3)
    d_inputs = scan_hop_fwd(d_outputs, coeffs_bwd, reverse=(not reverse))
    d_coeffs = torch.einsum('btci,btck->btcik', d_inputs,
                            shift(outputs, 1 if not reverse else -1))
    return d_inputs, d_coeffs
\end{lstlisting}
\vspace{4pt}
\noindent\textbf{(iv) Triton kernel for $m=2$.}
\begin{lstlisting}
@triton.jit
def _combine2(l0, l1, l2, l3, lb0, lb1, r0, r1, r2, r3, rb0, rb1):
    # compose two 2x2 affine maps: (A_r @ A_l, A_r @ b_l + b_r)
    return (r0 * l0 + r1 * l2, r0 * l1 + r1 * l3,
            r2 * l0 + r3 * l2, r2 * l1 + r3 * l3,
            r0 * lb0 + r1 * lb1 + rb0, r2 * lb0 + r3 * lb1 + rb1)

@triton.jit
def _slab_scan2_kernel(A_ptr, b_ptr, y_ptr, T, SLAB: tl.constexpr):
    # one program per row; each pass resolves SLAB time steps in log depth
    row = tl.program_id(0)
    s = tl.arange(0, SLAB)
    h0 = 0.0; h1 = 0.0
    for t0 in range(0, T, SLAB):
        t = t0 + s
        msk = t < T
        ao = row.to(tl.int64) * T * 4 + t.to(tl.int64) * 4
        bo = row.to(tl.int64) * T * 2 + t.to(tl.int64) * 2
        # masked lanes load the identity map so the scan is exact past the end
        a0 = tl.load(A_ptr + ao + 0, mask=msk, other=1.0).to(tl.float32)
        a1 = tl.load(A_ptr + ao + 1, mask=msk, other=0.0).to(tl.float32)
        a2 = tl.load(A_ptr + ao + 2, mask=msk, other=0.0).to(tl.float32)
        a3 = tl.load(A_ptr + ao + 3, mask=msk, other=1.0).to(tl.float32)
        c0 = tl.load(b_ptr + bo + 0, mask=msk, other=0.0).to(tl.float32)
        c1 = tl.load(b_ptr + bo + 1, mask=msk, other=0.0).to(tl.float32)
        p0, p1, p2, p3, y0, y1 = tl.associative_scan(
            (a0, a1, a2, a3, c0, c1), 0, _combine2)
        # fold in the state carried from earlier slabs
        y0 = y0 + p0 * h0 + p1 * h1
        y1 = y1 + p2 * h0 + p3 * h1
        tl.store(y_ptr + bo + 0, y0.to(y_ptr.dtype.element_ty), mask=msk)
        tl.store(y_ptr + bo + 1, y1.to(y_ptr.dtype.element_ty), mask=msk)
        last = s == SLAB - 1
        h0 = tl.sum(tl.where(last, y0, 0.0), 0)
        h1 = tl.sum(tl.where(last, y1, 0.0), 0)
\end{lstlisting}
\caption{Core of the block-diagonal scan implementations. Top: the PyTorch block hopscan (ii), whose composition operator is Eq.~\ref{eq:hopscan} and whose backward pass is the same scan run in reverse. Bottom: the Triton kernel (iv) for $m=2$; the four entries of $\mathbf{A}_t$ and the two of $\mathbf{b}_t$ are separate inputs of \texttt{tl.associative\_scan}, so the block product never leaves registers. The $m=4$ kernel is generated in the same way.}
\label{fig:a_code}
\end{figure}

\clearpage
\section{Proof of Proposition 1.}
\label{a:proof}

First, note that stability directly follows
from the induced norm bound.
We can reason blockwise: assuming $\sum_j|(\mathcal{A}^k_t)_{i,j}|\le 1$ implies that the eigenvalues of state-transition matrix $|\lambda^k_{i,t}|\le 1$.
Therefore, the product of such matrices will result in dynamical stability.

Next, by block-diagonality, it is sufficient to show that for all $k\in[1,N]$, $\|\mathbf{h}^k_{T}\|_\infty \le \max_{t\in [0,T]} \|\mathbf{v}^k_t\|_\infty$.
Let $h_{i,t}^k$ be the $i$-th coordinate of the generic $k$-th block hidden state $\mathbf{h}_t^k$ at time $t$.

\begin{equation}
\begin{bmatrix} h_{1,t}^k \\ \vdots \\ h_{m,t}^k \end{bmatrix} = \begin{bmatrix} 
a^{k}_{1,1,t} & \cdots & a^{k}_{1,m-1,t} & a^{k}_{1,m,t} \\
a^{k}_{2,1,t} & \cdots & a^{k}_{2,m-1,t} & a^{k}_{2,m,t} \\
\vdots & \ddots & \vdots & \vdots \\
a^{k}_{m,1,t} & \cdots & a^{k}_{m,m-1,t} & a^{k}_{m,m,t}\\ 
\end{bmatrix} \times \begin{bmatrix} h_{1,t-1}^k \\ \vdots \\ h_{m,t-1}^k \end{bmatrix} + \begin{bmatrix} a^{k}_{1,0,t} \\ \vdots \\ a^{k}_{m,0,t} \end{bmatrix} \odot \begin{bmatrix} v_{1,t}^k \\ \vdots \\  v_{m,t}^k \end{bmatrix}. 
\end{equation}

Hence, 
\begin{equation}
    h_{i,t}^k = \sum_{j=1}^m a^k_{i,j,t} h_{j,t-1}^k + a^k_{i,0,t}v^k_{i,t}.
    \label{eq_proof_1}
\end{equation}

It is then clear that by subadditivity of the absolute value,

\begin{equation}
    |h_{i,t}^k| \le \sum_{j=1}^m |a^k_{i,j,t}| \cdot |h_{j,t-1}^k| + |a^k_{i,0,t}| \cdot |v^k_{i,t}|.
\end{equation}

Hence, by collecting the non-coefficient terms, we find a further upper bound
\begin{equation}
    |h_{i,t}^k| \le \left(\sum_{j=1}^m |a^k_{i,j,t}| + |a^k_{i,0,t}| \right)\cdot \max\left[|v^k_{i,t}|,\max_{j\in[1,m]} |h_{j,t-1}^k| \right].
\end{equation}

By hypothesis, $\sum_{j=1}^m |a^k_{i,j,t}| + |a^k_{i,0,t}| = \sum_j|(\mathcal{A}^k_t)_{i,j}|\le 1$, and hence we conclude that

\begin{equation}
    |h_{i,t}^k| \le \max\left[|v^k_{i,t}|,\max_{j\in[1,m]} |h_{j,t-1}^k| \right].
    \label{eq:midproof}
\end{equation}

At this point, we can finalize the proof by induction.
We want to show that $\|\mathbf{h}^k_{T}\|_\infty \le \max_{t\in [0,T]} \|\mathbf{v}^k_t\|_\infty$.
Let us start from $T=1$.
Since $h_{i,0}^k = 0$ for all $i\in[1,m]$, we have

\begin{equation}
    h_{i,1}^k = a^k_{i,0,t}v^k_{i,1},
\end{equation}

hence, again because $\sum_j|(\mathcal{A}^k_0)_{i,j}|\le 1$, $|h_{i,1}^k| \le |v^k_{i,1}|$, we can conclude that $\|\mathbf{h}^k_{1}\|_\infty \le \|\mathbf{v}^k_1\|_\infty$.
Let us then assume by induction that $\|\mathbf{h}^k_{T-1}\|_\infty \le \max_{t\in [0,T-1]} \|\mathbf{v}^k_t\|_\infty$.
Recall that by Equation~\ref{eq:midproof},
\begin{align}
    |h_{i,t}^k| &\le \max\left[|v^k_{i,t}|,\max_{j\in[1,m]} |h_{j,t-1}^k| \right]\\
    &= \max\left[|v^k_{i,t}|,\|\mathbf{h}_{t-1}^k\|_\infty \right].
\end{align}

Hence,
\begin{align}
    \|\mathbf{h}_{t}^k\|_\infty &= \max_{j\in[1,m]} |h_{j,t}^k| \\
    &\le  \max_{j\in[1,m]} \max\left[|v^k_{i,t}|,\|\mathbf{h}_{t-1}^k\|_\infty \right]\\
     &=   \max\left[\max_{j\in[1,m]}|v^k_{i,t}|,\|\mathbf{h}_{t-1}^k\|_\infty \right]\\
     &\le   \max\left[\|\mathbf{v}^k_t\|_\infty,\max_{t\in [0,T-1]} \|\mathbf{v}^k_t\|_\infty\| \right]\\   
     &= \max_{t\in [0,T]} \|\mathbf{v}^k_t\|_\infty,
\end{align}
where in the second-last line we used the induction hypothesis.

\begin{corollary}
\label{cor:spec}
Under the normalization of Eq.~\ref{eq:norm}, every eigenvalue $\lambda$ of a block $\mathbf{A}^k_t$ satisfies $|\lambda|\le \max_i (1-|a^k_{i,0,t}|)$ for BD-LRU and $|\lambda|\le (1-|a^k_{0,t}|)^{1/m}$ for H-LRU, and the second bound is tight.
\end{corollary}
\noindent\textit{Proof.}
For BD-LRU, by the Gershgorin circle theorem every eigenvalue lies in a disc $|\lambda - a^k_{i,i,t}|\le\sum_{l\ne i}|a^k_{i,l,t}|$ for some row $i$, hence $|\lambda|\le\sum_{l=1}^m|a^k_{i,l,t}| = 1-|a^k_{i,0,t}|$.

For H-LRU (we omit the block index $k$), let $|\mathbf{A}_t|$ denote the entrywise absolute value of $\mathbf{A}_t$, which is the companion matrix of the nonnegative coefficients $|a_{1,t}|,\dots,|a_{m,t}|$.
By Wielandt's theorem~\citep[Thm.~8.1.18]{horn2012matrix}, $\rho(\mathbf{A}_t)\le\rho(|\mathbf{A}_t|)$, and by the Perron--Frobenius theorem $r:=\rho(|\mathbf{A}_t|)$ is itself an eigenvalue of $|\mathbf{A}_t|$, i.e., a nonnegative root of its characteristic polynomial, $r^m=\sum_{i=1}^m|a_{i,t}|\,r^{m-i}$.
Since the rows of $|\mathbf{A}_t|$ have absolute sum at most one, $r\le1$, so $r^{m-i}\le1$ for all $i$ and therefore $r^m\le\sum_{i=1}^m|a_{i,t}| = 1-|a_{0,t}|$, which gives $|\lambda|\le r\le(1-|a_{0,t}|)^{1/m}$.
Equality holds when $a_{1,t}=\dots=a_{m-1,t}=0$ and $|a_{m,t}|=1-|a_{0,t}|$, since then $\mathbf{A}_t^m=a_{m,t}\mathbf{I}$.
In addition, the determinant of the companion matrix gives $\prod_{i=1}^m|\lambda_i|=|a_{m,t}| = 1-\sum_{l=0}^{m-1}|a_{l,t}|$, so the input gate also bounds the product of the eigenvalue moduli. $\square$

\clearpage
\section{Eigenvalue analysis}
\label{a:eigen}

\begin{figure}[h]
     \centering
     \begin{subfigure}[b]{0.45\textwidth}
         \centering
        \includegraphics[width=\linewidth]{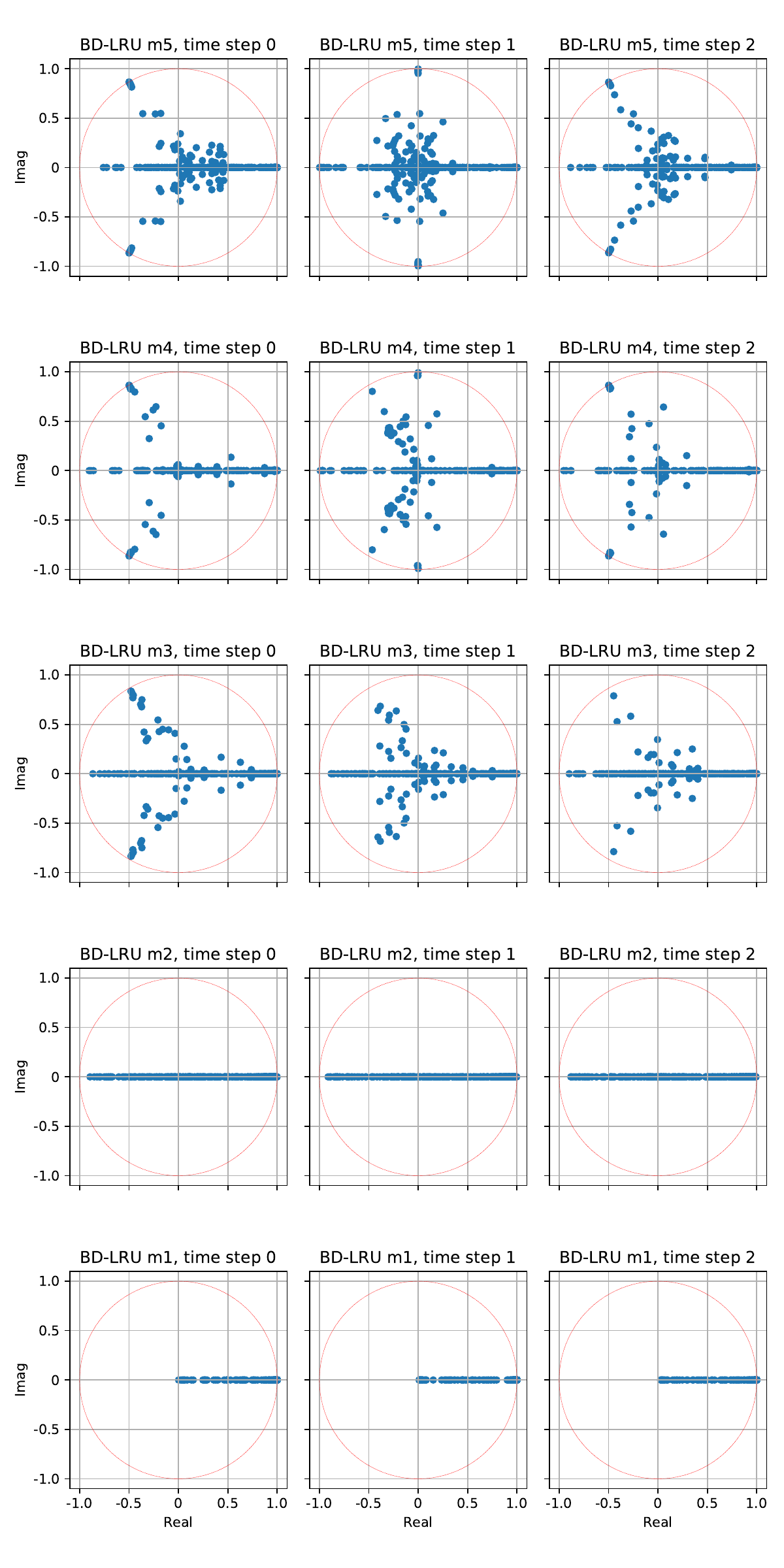}
        \caption{}
        \label{fig:eigens_bdlru}
     \end{subfigure}
     \hfill
     \begin{subfigure}[b]{0.45\textwidth}
         \centering
        \includegraphics[width=\linewidth]{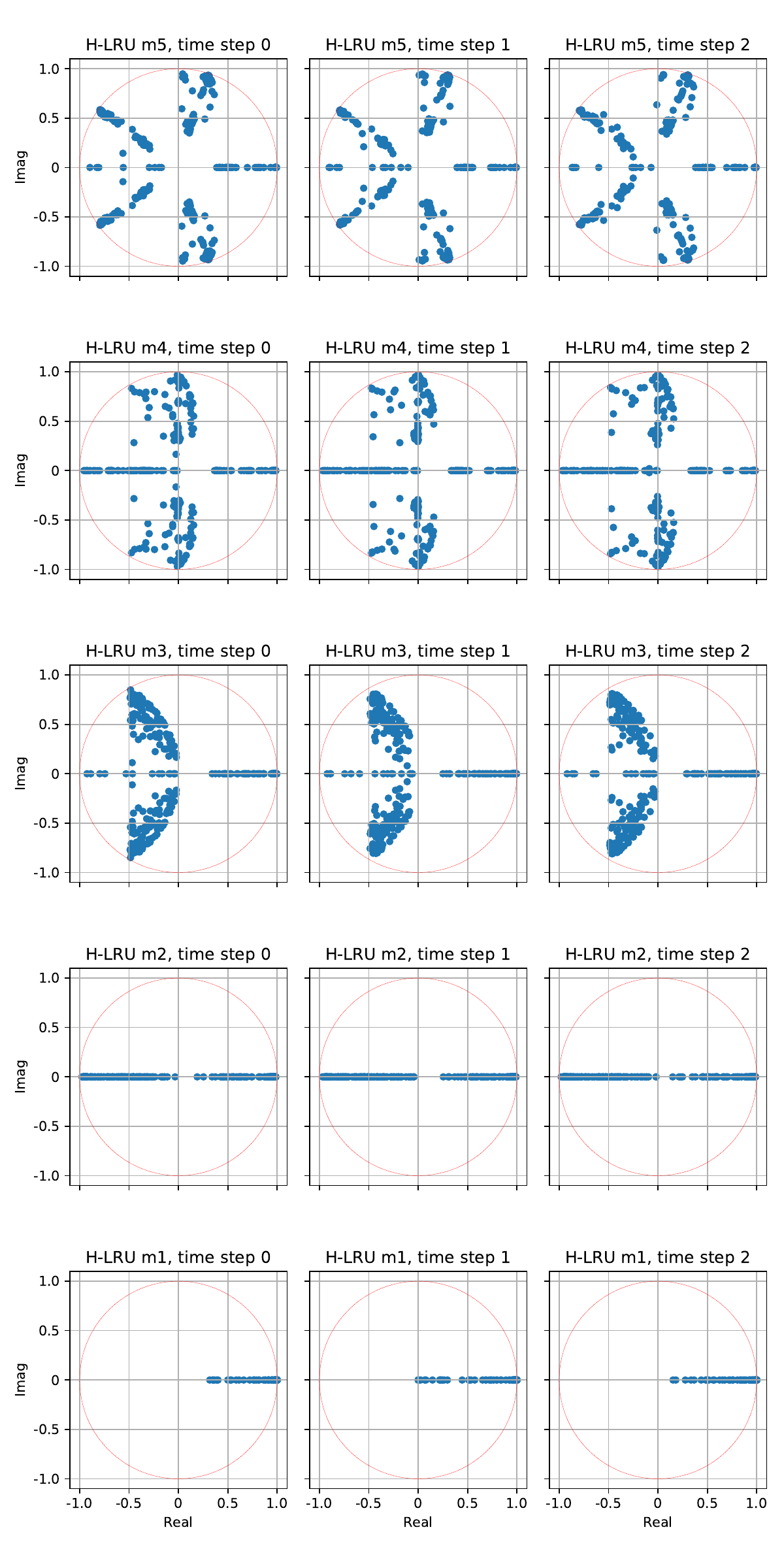}
        \caption{}
        \label{fig:eigens_hlru}
     \end{subfigure}

        \caption{Eigenvalues of
transition matrices in H-LRU/BD-LRU on $S5$ dataset. (a) BD-LRU with softmax normalization. (b) H-LRU with softmax normalization.
Each subplot corresponds to a specific time step (horizontal axis) and block size (vertical axis).
Note that as block size increases, the number of available symmetries increases as well. }
        \label{fig:eigen}
\end{figure}

\newpage
\section{Extended tables and additional figures}
\label{a:res}

\begin{table}[htbp]
\centering
\begin{tabular}{lrrrrrr}
\toprule
Models & $S_3$ (10k samples) & $S_4$ (50k) & $S_5$ (100k) \\
\midrule
\midrule
LSTM & \textbf{1.000 (1.000 $\pm$ 0.000)} & \textbf{1.000 (1.000 $\pm$ 0.000)} & \textbf{1.000 (1.000 $\pm$ 0.000)} \\
\midrule
Mamba2 c4 s2 & 0.640 (0.622 $\pm$ 0.020) & 0.423 (0.402 $\pm$ 0.033) & 0.268 (0.242 $\pm$ 0.028) \\
Mamba2 c4 s4 & 0.626 (0.609 $\pm$ 0.022) & 0.432 (0.396 $\pm$ 0.035) & 0.278 (0.237 $\pm$ 0.042) \\
Mamba2 c4 s8 & 0.657 (0.597 $\pm$ 0.055) & 0.429 (0.400 $\pm$ 0.025) & 0.261 (0.239 $\pm$ 0.033) \\
\midrule
DeltaNet[-1,1] nh8 & \textbf{1.000 (0.752 $\pm$ 0.223)} & 0.507 (0.456 $\pm$ 0.061) & 0.140 (0.136 $\pm$ 0.004) \\
\midrule
DeltaProduct$_4$[-1,1] nh8 & \textbf{1.000 (1.000 $\pm$ 0.000)} & \textbf{1.000 (0.868 $\pm$ 0.132)} & 0.140 (0.137 $\pm$ 0.004) \\
\midrule
PD-SSM & 1.000 (1.000 $\pm$ 0.000) & 0.569 (0.526 $\pm$ 0.037) & 0.347 (0.304 $\pm$ 0.045) \\
\midrule
BD-LRU m1 & 0.563 (0.563 $\pm$ 0.000) & 0.343 (0.323 $\pm$ 0.020) & 0.215 (0.199 $\pm$ 0.018) \\
BD-LRU m2 & \textbf{1.000 (1.000 $\pm$ 0.000)} & 0.699 (0.651 $\pm$ 0.067) & 0.338 (0.277 $\pm$ 0.057) \\
BD-LRU m3 & \textbf{1.000 (1.000 $\pm$ 0.000)} & \textbf{1.000 (0.930 $\pm$ 0.060)} & 0.484 (0.346 $\pm$ 0.148) \\
BD-LRU m4 & \textbf{1.000 (1.000 $\pm$ 0.000)} & \textbf{1.000 (0.969 $\pm$ 0.031)} & 0.882 (0.531 $\pm$ 0.308) \\
BD-LRU m5 & \textbf{1.000 (1.000 $\pm$ 0.000)} & \textbf{1.000 (0.914 $\pm$ 0.086)} & \textbf{1.000 (0.652 $\pm$ 0.302)} \\
BD-LRU m6 & \textbf{1.000 (1.000 $\pm$ 0.000)} & \textbf{1.000 (1.000 $\pm$ 0.000)} & \textbf{1.000 (0.923 $\pm$ 0.077)} \\
BD-LRU m8 & \textbf{1.000 (1.000 $\pm$ 0.000)} & \textbf{1.000 (1.000 $\pm$ 0.000)} & \textbf{1.000 (0.821 $\pm$ 0.176)} \\
BD-LRU m16 & \textbf{1.000 (1.000 $\pm$ 0.000)} & \textbf{1.000 (1.000 $\pm$ 0.000)} & \textbf{1.000 (1.000 $\pm$ 0.000)} \\
\midrule
H-LRU m1 & 0.573 (0.559 $\pm$ 0.013) & 0.347 (0.329 $\pm$ 0.016) & 0.230 (0.194 $\pm$ 0.045) \\
H-LRU m2 & 0.597 (0.583 $\pm$ 0.012) & 0.367 (0.354 $\pm$ 0.015) & 0.260 (0.236 $\pm$ 0.031) \\
H-LRU m3 & 0.610 (0.589 $\pm$ 0.030) & 0.399 (0.381 $\pm$ 0.016) & 0.321 (0.301 $\pm$ 0.033) \\
H-LRU m4 & 0.624 (0.604 $\pm$ 0.018) & 0.414 (0.390 $\pm$ 0.021) & 0.337 (0.307 $\pm$ 0.031) \\
H-LRU m5 & 0.622 (0.609 $\pm$ 0.019) & 0.448 (0.424 $\pm$ 0.025) & 0.377 (0.337 $\pm$ 0.065) \\
H-LRU m6 & 0.630 (0.596 $\pm$ 0.053) & 0.446 (0.431 $\pm$ 0.016) & 0.386 (0.311 $\pm$ 0.065) \\
H-LRU m8 & 0.641 (0.612 $\pm$ 0.029) & 0.494 (0.451 $\pm$ 0.037) & 0.390 (0.351 $\pm$ 0.062) \\
H-LRU m16 & 0.659 (0.583 $\pm$ 0.066) & 0.512 (0.459 $\pm$ 0.065) & 0.393 (0.331 $\pm$ 0.063) \\
\bottomrule
\bottomrule
\end{tabular}
\caption{Model performance on permutation composition tasks for different datasets of different sizes: $S_3$ (10k training samples), $S_4$ (50k training samples), $S_5$ (100k training samples).
The accuracy values reflect the impact of window size (H-LRU) and block size (BD-LRU), both denoted by $m$.
We note that BD-LRU performance improves with block size, demonstrating strong sample efficiency by solving the tasks even given limited training data.
All models are single-layer configurations with a maximum overall hidden dimension of 6144.
The presented results are the best test accuracies (mean $\pm$ std in brackets across 5 seeds).
DeltaProduct and DeltaNet are reported with 8 heads (nh) and [-1,1] denotes the range of eigenvalues.
Mamba2 are reported with convolution size (c) and state space extension (s).
We utilized all permutations belonging to the symmetric groups $S_3$, $S_4$, $S_5$ (6, 24, and 120 elements, respectively).}
\label{tab:s_results_ext}
\end{table}

\begin{table}[htbp]
\centering
\begin{tabular}{lll}
\toprule
Models & S3 (250) & S4 (3k) \\
\midrule
LSTM & 0.316 (0.298 $\pm$ 0.016) & 0.368 (0.297 $\pm$ 0.093) \\
\midrule
Mamba2 c4 s2 & 0.284 (0.277 $\pm$ 0.006) & 0.115 (0.113 $\pm$ 0.003) \\
Mamba2 c4 s4 & 0.284 (0.278 $\pm$ 0.006) & 0.115 (0.105 $\pm$ 0.009) \\
Mamba2 c4 s8 & 0.281 (0.279 $\pm$ 0.002) & 0.119 (0.115 $\pm$ 0.004) \\
\midrule
DeltaNet[-1,1] nh8 & 0.264 (0.259 $\pm$ 0.005) & 0.138 (0.117 $\pm$ 0.018) \\
\midrule
DeltaProduct$_4$[-1,1] nh8 & 0.279 (0.273 $\pm$ 0.008) & 0.136 (0.136 $\pm$ 0.000) \\
\midrule
PD-SSM & 0.273 (0.270 $\pm$ 0.005) & 0.154 (0.125 $\pm$ 0.025) \\
\midrule
BD-LRU m1 & 0.381 (0.347 $\pm$ 0.034) & 0.218 (0.204 $\pm$ 0.012) \\
BD-LRU m2 & 0.488 (0.459 $\pm$ 0.032) & 0.356 (0.265 $\pm$ 0.079) \\
BD-LRU m3 & \textbf{1.000 (0.998 $\pm$ 0.002)} & 0.426 (0.382 $\pm$ 0.038) \\
BD-LRU m4 & \textbf{1.000 (0.792 $\pm$ 0.208)} & \textbf{1.000 (0.781 $\pm$ 0.219)} \\
BD-LRU m5 & \textbf{1.000 (0.812 $\pm$ 0.188)} & \textbf{1.000 (0.582 $\pm$ 0.363)} \\
BD-LRU m6 & \textbf{1.000 (1.000 $\pm$ 0.000)} & \textbf{1.000 (0.849 $\pm$ 0.151)} \\
BD-LRU m8 & 0.995 (0.962 $\pm$ 0.038) & \textbf{1.000 (0.737 $\pm$ 0.263)} \\
BD-LRU m16 & \textbf{1.000 (0.959 $\pm$ 0.041)} & \textbf{1.000 (0.792 $\pm$ 0.208)} \\
\midrule
H-LRU m1 & 0.363 (0.351 $\pm$ 0.014) & 0.212 (0.203 $\pm$ 0.008) \\
H-LRU m2 & 0.311 (0.310 $\pm$ 0.001) & 0.194 (0.182 $\pm$ 0.016) \\
H-LRU m3 & 0.316 (0.298 $\pm$ 0.016) & 0.206 (0.179 $\pm$ 0.023) \\
H-LRU m4 & 0.309 (0.292 $\pm$ 0.015) & 0.191 (0.175 $\pm$ 0.013) \\
H-LRU m5 & 0.318 (0.300 $\pm$ 0.015) & 0.194 (0.179 $\pm$ 0.014) \\
H-LRU m6 & 0.283 (0.281 $\pm$ 0.003) & 0.168 (0.165 $\pm$ 0.003) \\
H-LRU m8 & 0.281 (0.276 $\pm$ 0.004) & 0.166 (0.164 $\pm$ 0.002) \\
H-LRU m16 & 0.260 (0.255 $\pm$ 0.004) & 0.159 (0.157 $\pm$ 0.001) \\
\bottomrule
\bottomrule
\end{tabular}
\caption{Extension of Table \ref{tab:s_results_ext} for permutation composition tasks that have limited number of training samples:  $S_3$ (250 training samples), $S_4$ (3k training samples).
We note that BD-LRU performance improves with block size, demonstrating strong sample efficiency by solving the tasks even given limited training data.}
\label{tab:s_results_ext_eff}
\end{table}

\renewcommand{\arraystretch}{0.9}
\begin{table}[htbp]
\centering
\begin{tabular}{lrrrrrrrrrrrr}
\toprule
Models & In-Context Recall Tasks & Selective Copy Tasks  & Compression Tasks \\
\midrule
\midrule
LSTM & \textbf{1.000 (1.000 $\pm$ 0.000)} & \textbf{1.000 (0.998 $\pm$ 0.002)} & 0.749 (0.697 $\pm$ 0.092) \\
\midrule
Mamba2 c2 s2 & \textbf{1.000 (0.970 $\pm$ 0.030)} & 0.791 (0.743 $\pm$ 0.051) & 0.623 (0.508 $\pm$ 0.142) \\
Mamba2 c2 s4 & \textbf{1.000 (0.988 $\pm$ 0.012)} & 0.812 (0.786 $\pm$ 0.032) & 0.721 (0.582 $\pm$ 0.131) \\
Mamba2 c2 s8 & \textbf{1.000 (1.000 $\pm$ 0.000)} & 0.830 (0.799 $\pm$ 0.032) & 0.606 (0.535 $\pm$ 0.069) \\
Mamba2 c4 s2 & \textbf{1.000 (1.000 $\pm$ 0.000)} & 0.798 (0.753 $\pm$ 0.059) & 0.623 (0.528 $\pm$ 0.108) \\
Mamba2 c4 s4 & \textbf{1.000 (1.000 $\pm$ 0.000)} & 0.800 (0.756 $\pm$ 0.060) & 0.637 (0.548 $\pm$ 0.098) \\
Mamba2 c4 s8 & \textbf{1.000 (1.000 $\pm$ 0.000)} & 0.803 (0.742 $\pm$ 0.073) & 0.634 (0.561 $\pm$ 0.065) \\
\midrule
DeltaNet[-1,1]  nh2 & \textbf{1.000 (1.000 $\pm$ 0.000)} & 0.796 (0.744 $\pm$ 0.045) & 0.709 (0.470 $\pm$ 0.180) \\
DeltaNet[-1,1]  nh4 & \textbf{1.000 (1.000 $\pm$ 0.000)} & 0.879 (0.806 $\pm$ 0.065) & 0.718 (0.654 $\pm$ 0.068) \\
DeltaNet[-1,1]  nh8 & \textbf{1.000 (1.000 $\pm$ 0.000)} & 0.902 (0.870 $\pm$ 0.040) & 0.765 (0.665 $\pm$ 0.110) \\
\midrule
DeltaProduct$_4$[-1,1] nh2 & \textbf{1.000 (1.000 $\pm$ 0.000)} & 0.996 (0.985 $\pm$ 0.011) & 0.753 (0.648 $\pm$ 0.090) \\
DeltaProduct$_4$[-1,1] nh4 & \textbf{1.000 (1.000 $\pm$ 0.000)} & \textbf{1.000 (0.996 $\pm$ 0.004)} & 0.720 (0.645 $\pm$ 0.096) \\
DeltaProduct$_4$[-1,1] nh8 & \textbf{1.000 (1.000 $\pm$ 0.000)} & \textbf{1.000 (0.998 $\pm$ 0.001)} & 0.735 (0.655 $\pm$ 0.073) \\
\midrule
PD-SSM & 0.817 (0.785 $\pm$ 0.012) & 0.446 (0.316 $\pm$ 0.112) & 0.338 (0.299 $\pm$ 0.020) \\
\midrule
BD-LRU m1 & 0.779 (0.721 $\pm$ 0.036) & 0.835 (0.815 $\pm$ 0.016) & 0.725 (0.580 $\pm$ 0.119) \\
BD-LRU m2 & \textbf{1.000 (1.000 $\pm$ 0.000)} & 0.964 (0.954 $\pm$ 0.009) & 0.760 (0.645 $\pm$ 0.129) \\
BD-LRU m3 & \textbf{1.000 (1.000 $\pm$ 0.000)} & 0.984 (0.975 $\pm$ 0.006) & 0.764 (0.642 $\pm$ 0.126) \\
BD-LRU m4 & \textbf{1.000 (1.000 $\pm$ 0.000)} & 0.985 (0.976 $\pm$ 0.006) & \textbf{0.786 (0.748 $\pm$ 0.047)} \\
BD-LRU m5 & \textbf{1.000 (1.000 $\pm$ 0.000)} & 0.985 (0.978 $\pm$ 0.005) & 0.782 (0.732 $\pm$ 0.056) \\
BD-LRU m6 & \textbf{1.000 (1.000 $\pm$ 0.000)} & 0.983 (0.977 $\pm$ 0.004) & 0.775 (0.665 $\pm$ 0.101) \\
BD-LRU m8 & \textbf{1.000 (1.000 $\pm$ 0.000)} & 0.996 (0.946 $\pm$ 0.032) & 0.748 (0.685 $\pm$ 0.079) \\
BD-LRU m16 & \textbf{1.000 (1.000 $\pm$ 0.000)} & 0.999 (0.991 $\pm$ 0.009) & 0.725 (0.635 $\pm$ 0.121) \\
\midrule
H-LRU m1 & 0.783 (0.721 $\pm$ 0.038) & 0.847 (0.813 $\pm$ 0.024) & 0.758 (0.650 $\pm$ 0.133) \\
H-LRU m2 & 0.998 (0.993 $\pm$ 0.004) & 0.853 (0.827 $\pm$ 0.023) & 0.772 (0.692 $\pm$ 0.061) \\
H-LRU m3 & \textbf{1.000 (1.000 $\pm$ 0.000)} & 0.853 (0.830 $\pm$ 0.018) & 0.773 (0.673 $\pm$ 0.136) \\
H-LRU m4 & \textbf{1.000 (1.000 $\pm$ 0.000)} & 0.845 (0.823 $\pm$ 0.019) & 0.774 (0.736 $\pm$ 0.038) \\
H-LRU m5 & \textbf{1.000 (1.000 $\pm$ 0.000)} & 0.835 (0.823 $\pm$ 0.009) & 0.776 (0.698 $\pm$ 0.070) \\
H-LRU m6 & \textbf{1.000 (0.954 $\pm$ 0.026)} & 0.817 (0.805 $\pm$ 0.009) & 0.775 (0.674 $\pm$ 0.129) \\
H-LRU m8 & \textbf{1.000 (0.976 $\pm$ 0.024)} & 0.807 (0.776 $\pm$ 0.022) & 0.766 (0.694 $\pm$ 0.060) \\
H-LRU m16 & \textbf{1.000 (0.996 $\pm$ 0.003)} & 0.679 (0.611 $\pm$ 0.044) & 0.708 (0.610 $\pm$ 0.114) \\
\bottomrule
\bottomrule
\end{tabular}
\caption{Performance on the in-context recall, selective copying and compression tasks.
The presented results are the average of best test accuracies (mean $\pm$ std in brackets across 5 seeds) across four configurations of the corresponding synthetic dataset with different vocabulary sizes, sequence lengths and number of training examples.
Results are shown for different window (H-LRU) and block sizes (BD-LRU) $m$.
DeltaProduct and DeltaNet are reported for different number of heads (nh) and [-1,1] denotes the range of eigenvalues.
Mamba2 are reported variants are reported with different convolution sizes (c) and state space extension (s).
Note that overall performance of our models consistently improves with window/block size up to approximately 3–5, after which the gains saturate or exhibit slight degradation.
All models are single-layer configurations with a maximum overall hidden dimension of 6144. }
\label{tab:mad_results_ext}
\end{table}

\begin{table}[htbp]
\centering
\begin{tabular}{lrrr}
\toprule
Models & Memorization Tasks & Fuzzy Recall Tasks & Noisy Recall Tasks \\
\midrule
\midrule
LSTM & 0.787 (0.778 $\pm$ 0.006) & 0.128 (0.120 $\pm$ 0.005) & 0.999 (0.999 $\pm$ 0.000) \\
\midrule
PD-SSM & 0.787 (0.777 $\pm$ 0.006) & 0.117 (0.109 $\pm$ 0.003) & 0.800 (0.750 $\pm$ 0.027) \\
\midrule
DeltaNet[-1,1] & 0.357 (0.339 $\pm$ 0.013) & \textbf{0.525 (0.461 $\pm$ 0.035)} & \textbf{1.000 (1.000 $\pm$ 0.000)} \\
\midrule
DeltaProduct$_4$[-1,1] & 0.332 (0.316 $\pm$ 0.003) & 0.373 (0.309 $\pm$ 0.045) & \textbf{1.000 (1.000 $\pm$ 0.000)} \\
\midrule
BD-LRU m1 & 0.785 (0.777 $\pm$ 0.005) & 0.142 (0.136 $\pm$ 0.002) & 0.687 (0.681 $\pm$ 0.006) \\
BD-LRU m2 & 0.785 (0.776 $\pm$ 0.005) & 0.293 (0.267 $\pm$ 0.013) & \textbf{1.000 (1.000 $\pm$ 0.000)} \\
BD-LRU m4 & 0.787 (0.776 $\pm$ 0.006) & 0.252 (0.241 $\pm$ 0.004) & \textbf{1.000 (1.000 $\pm$ 0.000)} \\
BD-LRU m8 & 0.785 (0.776 $\pm$ 0.005) & 0.177 (0.165 $\pm$ 0.009) & \textbf{1.000 (0.999 $\pm$ 0.000)} \\
BD-LRU m16 & 0.785 (0.776 $\pm$ 0.005) & 0.135 (0.130 $\pm$ 0.003) & 0.986 (0.980 $\pm$ 0.007) \\
\midrule
H-LRU m1 & 0.786 (0.776 $\pm$ 0.005) & 0.138 (0.132 $\pm$ 0.005) & 0.689 (0.685 $\pm$ 0.001) \\
H-LRU m2 & \textbf{0.789 (0.780 $\pm$ 0.006)} & 0.175 (0.156 $\pm$ 0.009) & \textbf{1.000 (1.000 $\pm$ 0.000)} \\
H-LRU m4 & 0.784 (0.774 $\pm$ 0.006) & 0.174 (0.153 $\pm$ 0.010) & \textbf{1.000 (1.000 $\pm$ 0.000)} \\
H-LRU m8 & 0.753 (0.744 $\pm$ 0.005) & 0.168 (0.152 $\pm$ 0.007) & 0.910 (0.886 $\pm$ 0.011) \\
H-LRU m16 & 0.658 (0.650 $\pm$ 0.003) & 0.169 (0.154 $\pm$ 0.004) & 0.828 (0.806 $\pm$ 0.010) \\
\bottomrule
\bottomrule
\end{tabular}
\caption{Performance on the memorization, fuzzy in-context recall and noisy in-context recall tasks of the MAD benchmark, which are not part of the aggregate scores reported in the main text.
The presented results are the average of best test accuracies (mean $\pm$ std in brackets across 3 seeds) across three configurations of the memorization task with different vocabulary sizes (2048, 4096 and 8192) and four configurations of each recall task with different vocabulary sizes, sequence lengths and numbers of training examples.
All models are single-layer configurations with the sequence-mixing layer matched to $\sim$1M parameters.
Memorization accuracy is capped by the fraction of test keys that appear in the 256 training sequences (0.98, 0.86 and 0.64 for the three vocabulary sizes); LSTM, PD-SSM, BD-LRU at every block size and H-LRU up to $m=4$ all reach this ceiling, so the task does not separate them, while the delta-rule models fall far below it.
Noisy recall is solved by the delta-rule models and by BD-LRU and H-LRU at moderate block sizes, and fuzzy recall ranks the models in the same order as in-context recall (Table~\ref{tab:mad_results_ext}), with the delta-rule family ahead.
DeltaProduct is reported with four Householder factors and [-1,1] denotes the range of eigenvalues; PD-SSM uses a PyTorch associative scan.}
\label{tab:mad_extra}
\end{table}

\begin{figure}[htbp]
    \centering
    \includegraphics[width=\linewidth]{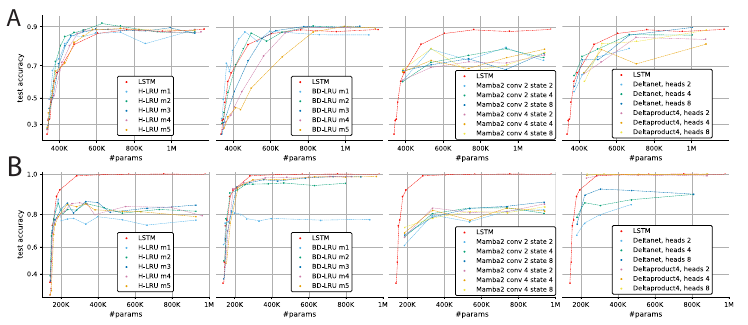}
    \caption{Performance of different single-layer models as a function of the hidden size (up to 6k overall hidden dimension) in the compression task (\textbf{A}) and the selective copying task (\textbf{B}).
Results are shown for different window sizes (H-LRU) and block sizes (BD-LRU) $m$.
We compare our networks with different configurations of Mamba2 (with two convolution kernel sizes (2,4) and several values of the state space expansion factor (2,4,8)).
For comparison to low-rank models, we also include DeltaNet and DeltaProduct with 4 Householder transforms with different number of heads (2,4,8).}
    \label{fig:mad_scaling}
\end{figure}

\clearpage
\section{Selectivity ablation}
\label{a:sel}

To isolate and quantify the contribution of selectivity, we conducted an ablation study.
In this analysis, the input-dependent selective gates in both the \ref{eq:hlru} and \ref{eq:bdlru} architectures were replaced with data-invariant, learnable parameters.

As hypothesized, the non-selective variants exhibited a significant performance degradation compared to their selective counterparts on our synthetic benchmark.
On tasks requiring dynamic token manipulation—such as in-context recall, selective copying, and permutation composition—the non-selective models failed to achieve meaningful performance.
For these tasks, increasing the window or block size yielded no discernible improvement, confirming the necessity of selectivity.

\begin{figure}[h]
    \centering
    \includegraphics[width=\linewidth]{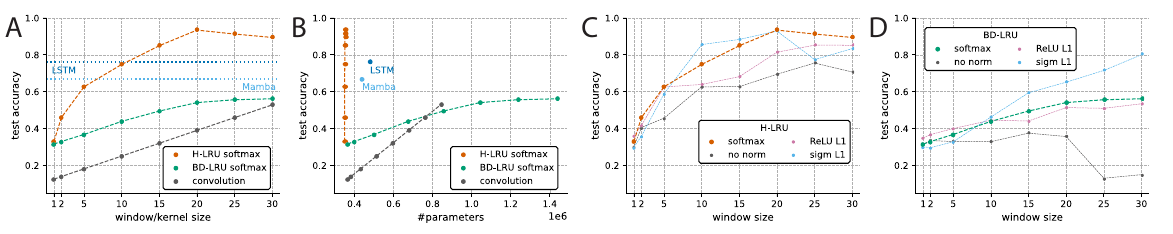}
    \caption{Scaling analysis of non-selective models on the compression task.
\textbf{A}.
Performance as a function of window size $m$ of non-selective higher-order LRU (H-LRU) and block size $m$ of block diagonal LRU (BD-LRU).
For the convolutional baseline, the performance presented as a function of kernel size.
\textbf{B}.
The same results plotted against parameter count.
Note that scaling with window size of non-selective H-LRU demonstrates extreme parameter efficiency, resulting in a nearly vertical trajectory on the plot.
\textbf{C}.
Comparison of scaling properties between different parameterizations for H-LRU.
\textbf{D}.
Comparison of scaling properties between different parameterizations for BD-LRU.
    }
    \label{fig:nonsel_norm}
\end{figure}

However, the results on the compression task were more nuanced, see Fig. \ref{fig:nonsel_norm}.
We observed that our proposed $L1$ normalization scheme enabled the non-selective models to improve with larger block and window sizes, albeit at a lower rate than their selective analogs. 

To highlight the advantages of recurrent architectures, we used a convolution layer as a baseline.
This model is limited to explicit, local time mixing within its kernel, in contrast to the implicit and unbounded temporal integration provided by a hidden state.
Our experiments showed that H-LRU decisively outperforms the convolution on the compression task.
This demonstrates the critical role of recurrent state mixing for tasks requiring efficient long-range temporal reasoning.
Furthermore, the non-selective H-LRU with large window sizes ($m>15$) demonstrated strong performance, surpassing the LSTM and Mamba2 baselines and even approaching the performance of our selective models.
This finding highlights the powerful inductive bias of the higher-order recurrence for parameter-efficient compression.

In contrast, the non-selective BD-LRU performed poorly on the compression task, only marginally surpassing the convolution baseline.
Interestingly, for this non-selective variant, the sigmoidal $L1$ normalization outperformed softmax normalization, highlighting a difference in how these schemes interact with selective versus fixed parameterizations. 

In addition, when we analyzed H-LRU with minimal point-wise selective gates that don't mix channel dimensions, we observed a very moderate improvement in compression task.
This indicates that not only selectivity itself but also density of selectivity in gates plays an important role in improving networks' expressivity.

While the overall performance of these non-selective models is modest, their parameter efficiency can become advantageous in resource-constrained settings.
Given the strong compression results of the non-selective H-LRU, we hypothesize that such models could be optimized for use as highly efficient embedding layers, a direction we leave for future research.

\clearpage
\section{Relation between the expressivity of LRUs and State Space Duality}

Recently, it has been shown that there is a direct correspondence between state space models, the Transformer architecture and structured attention matrices~\citep{dao2024transformers}.
Following this approach, we can reformulate the general LRU as a general discrete time SSM
\begin{equation}
\begin{aligned}
\mathbf{h}_{t} &= \mathbf{A}_t \times \mathbf{h}_{t-1} + \mathbf{B}_t \times \mathbf{v}_t\\
\mathbf{y}_t &= \mathbf{C}_t  \times \mathbf{h}_t.
\label{eq:ssm}
\end{aligned}
\end{equation}
Here, we consider the general case of SSMs in which mixing matrices $ \mathbf{C}_t, \mathbf{A}_t, \mathbf{B}_t$ are dense matrices.
We note that although state space models are commonly defined in continuous time, they have to be discretized for implementation, at which point they conform to the discrete form described by Eq.~\ref{eq:ssm}.
In this study, we effectively ignored the role of $\mathbf{C}_t$, but it can be introduced without affecting the validity of our arguments.
 
Following the approach to reformulating state space models (SSMs) as attention mechanisms, the architecture given in Eq.~\ref{eq:ssm} can be expressed in block matrix representation assuming a fixed sequence length $T$:
\begin{equation*}
\begin{bmatrix}
   \mathbf{y}_1 \\
   \mathbf{y}_2 \\
   \mathbf{y}_3 \\
   \vdots \\
   \mathbf{y}_T \\
   \end{bmatrix}
 =
  \begin{bmatrix}
   \mathbf{C}_1  \mathbf{B}_1 &
   \mathbf{0}  &
   \mathbf{0}  &
  \cdots &
   \mathbf{0} \\
   \mathbf{C}_2  \mathbf{A}_1 \mathbf{B}_1  &
  \mathbf{C}_2  \mathbf{B}_2 &
  \mathbf{0}  &
  \cdots &
   \mathbf{0}   \\
   \mathbf{C}_3  \mathbf{A}_2 \mathbf{A}_1 \mathbf{B}_1  &
  \mathbf{C}_3  \mathbf{A}_2 \mathbf{B}_2 &
  \mathbf{C}_3  \mathbf{B}_3 &
  \cdots &
   \mathbf{0}   \\
   \vdots &
   \vdots &
   \vdots &
   \ddots &
   \vdots  \\
   \mathbf{C}_T \prod_{j=1}^{T} \mathbf{A}_j \mathbf{B}_1  &
   \mathbf{C}_T \prod_{j=2}^{T} \mathbf{A}_j \mathbf{B}_2  &
    \cdots &
    \cdots &
  \mathbf{C}_T  \mathbf{B}_T \\
   \end{bmatrix} \begin{bmatrix}
   \mathbf{v}_1 \\
   \mathbf{v}_2 \\
   \mathbf{v}_3 \\
   \vdots \\
   \mathbf{v}_T \\
   \end{bmatrix}
   \label{eq:ssd_form}
\end{equation*}
If we abstract the details of SSMs matrices, we obtain the generalized attention formulation:
\begin{equation}
\begin{bmatrix}
   \mathbf{y}_1 \\
   \mathbf{y}_2 \\
   \mathbf{y}_3 \\
   \vdots \\
   \end{bmatrix}
 =
  \begin{bmatrix}
   \overline{\mathbf{A}}_{1,1} &
   \mathbf{0}  &
   \mathbf{0} &
  \cdots &
   \mathbf{0} \\
    \overline{\mathbf{A}}_{2,1} &
   \overline{\mathbf{A}}_{2,2}   &
   \mathbf{0} &
  \cdots &
   \mathbf{0}   \\
   \overline{\mathbf{A}}_{3,1} &
  \overline{\mathbf{A}}_{3,2} &
   \overline{\mathbf{A}}_{3,3} &
   \ddots &
   \vdots  \\
   \vdots &
   \vdots &
   \vdots &
   \ddots &
   \vdots  \\
   \end{bmatrix} \begin{bmatrix}
   \mathbf{v}_1 \\
   \mathbf{v}_2 \\
   \mathbf{v}_3 \\
   \vdots \\
   \end{bmatrix}.
   \label{eq:att_form}
\end{equation}
Importantly, elements $\overline{\mathbf{A}}_{k,l}$ of the block attention matrix are also matrices in this representation.
According to State Space Duality~\citep{dao2024transformers}, both the attention in Transformers and diagonal SSMs result in diagonal matrices $\overline{\mathbf{A}}_{k,l}$.
So, their architecture allows for efficient parallelization as it separates temporal mixing from channel mixing.

In contrast to diagonal SSMs and LRUs, both H-LRU and BD-LRU architectures result in block-diagonal matrices $\overline{\mathbf{A}}_{k,l}$, allowing richer but block-limited channel mixing inside the generalized block attention matrix in Eq.~\ref{eq:att_form}.
Such channel mixing allows for the state mixing patterns that are not accessible to one layer of diagonal LRU or SSMs.
Although the channel mixing in H-LRU is more expressive than the one in a diagonal LRU, it is still more restricted compared to BD-LRU (it is equivalent to mixing only in one row of a block-diagonal matrix), placing the expressivity of H-LRU between diagonal LRU and BD-LRU. 
Notably, if we extend SSMs with higher-order or block-diagonal structures, their expressivity would lag behind analogous LRUs due to the restrictions on mixing patterns imposed by the chosen discretization scheme.
Overall, the generalized block attention formulation~\ref{eq:att_form} reveals that diagonal, higher-order, block diagonal and dense variants of LRUs and SSMs form a hierarchy of architectures, each providing access to increasingly complex state mixing patterns which result in increased expressivity.

\clearpage
\section{Chomsky Hierarchy Tasks}
\label{a:chom}

The Chomsky hierarchy formalizes increasing levels of expressiveness and computational complexity of formal languages into several hierarchical classes \citep{chomsky1956three,deletang2022neural}.
Here, we tested several tasks from this hierarchy: Parity, Cycle Navigation, Modular Arithmetic with and without brackets.
The Parity task requires computing whether given binary string is even or not.
Cycle Navigation requires computing the end position given a sequence of movements on a cycle of length 5.
Modular Arithmetic tasks require computing the result modulo 5 for a given sequence of numbers in $(0, 1, 2, 3, 4)$ and operations in $(+,-,\cdot)$, with or without brackets.

In our experiments, we observe that similar to $S_3$ task, the Parity task can be solved by BD-LRU with access to negative eigenvalues ($m\geq2$).
For the Cycle Navigation task we obtain similar results as for $S_5$ task.
BD-LRU is able to solve it starting from $m=5$.
Therefore, the results on these two tasks from Chomsky Hierarchy support our previously found advantage of BD-LRUs on permutation tasks. 

Modular arithmetic tasks present a challenge for highly parallel Transformer architecture, often require grokking and have pure generalization \citep{gromov2023grokking}.
In contrast, it has been shown that the sequential nature of state mixing in RNNs has a strongly beneficial bias for arithmetic-like induction \citep{merrill2023parallelism}.
However, both our linear variants and other modern LRNNs struggle with such arithmetic tasks \citep{siems2025deltaproduct}, supporting the idea that the nonlinearity of state transitions is crucial in such tasks \citep{chang2024language}.
In our experiments, we found that BD-LRUs were able to solve Modular Arithmetic without brackets, while the version with brackets remained challenging, similar to other RNNs.
  
\begin{table}[h]
\centering
\begin{tabular}{lrrrrrrrrrrrr}
\toprule
Models & cycle nav & mod arith no brack & mod arith w brack & parity \\
\midrule
\midrule
LSTM & \textbf{1.000} & 0.976 & 0.663 & \textbf{1.000}   \\
\midrule
BD-LRU m1 & 0.434 & 0.370 & 0.370 & 0.512 \\
BD-LRU m2 & 0.425 & 0.493 & 0.417 & \textbf{1.000}  \\
BD-LRU m3 & 0.597 & 0.546 & 0.434 & \textbf{1.000}  \\
BD-LRU m4 & 0.608 & 0.459 & 0.435 & \textbf{1.000}  \\
BD-LRU m5 & \textbf{1.000}  & 0.525 & 0.422 & \textbf{1.000} \\
BD-LRU m6 & \textbf{1.000}  & 0.433 & 0.440 & \textbf{1.000} \\
BD-LRU m8 & \textbf{1.000}  & 0.553 & 0.395 & \textbf{1.000} \\
BD-LRU m16 & \textbf{1.000} & \textbf{1.000} & 0.448 & \textbf{1.000} \\
\bottomrule
\bottomrule
\end{tabular}
\caption{Accuracy on Chomsky hierarchy
tasks.}
\label{tab:ch_results}
\end{table}

\newpage

\newpage
\section*{NeurIPS Paper Checklist}

\begin{enumerate}

\item {\bf Claims}
    \item[] Question: Do the main claims made in the abstract and introduction accurately reflect the paper's contributions and scope?
    \item[] Answer: \answerYes{},%
    \item[] Justification: Yes. The abstract and introduction accurately reflect the paper's core contributions and scope. Specifically, they clearly outline the introduction of the H-LRU and BD-LRU architectures, the design of the selective gate normalization scheme, the provision of our parallel-scan implementation, and our empirical findings demonstrating that the structure of state mixing is a critical driver of performance. 

\item {\bf Limitations}
    \item[] Question: Does the paper discuss the limitations of the work performed by the authors?
    \item[] Answer: \answerYes{}
    \item[] Justification: Yes. In the Conclusion and Outlook section, we explicitly address limitations. We note that the optimal block or window size is inherently task-dependent, and we provide practical guidelines for practitioners to navigate this. As a potential limitation, we note that we do not evaluate our models on large-scale language modeling. We emphasize that we do not aim to establish a new SOTA on large-scale language modeling. Our primary objective is to study structured state mixing.

\item {\bf Theory assumptions and proofs}
    \item[] Question: For each theoretical result, does the paper provide the full set of assumptions and a complete (and correct) proof?
    \item[] Answer: \answerYes{}
    \item[] Justification: The main text contains one theoretical proposition, for which the complete proof and all underlying assumptions are fully detailed in the Appendix \ref{a:proof}.

    \item {\bf Experimental result reproducibility}
    \item[] Question: Does the paper fully disclose all the information needed to reproduce the main experimental results of the paper to the extent that it affects the main claims and/or conclusions of the paper (regardless of whether the code and data are provided or not)?
    \item[] Answer: \answerYes{}
    \item[] Justification: Yes. The paper fully discloses all necessary details to reproduce the main experimental results. The proposed model architectures are described in full mathematical and structural detail, and the implementation of our parallel scan is explicitly outlined. Hyperparameter configurations for both our proposed architectures and all baseline models are documented. Furthermore, complete descriptions of the dataset configurations are included in the appendix. In addition, we also provide direct references to the source code for both the synthetic \url{https://github.com/IgorDn8/mad-lab} and language modeling \url{https://github.com/IgorDn8/plainLM} experiments.

\item {\bf Open access to data and code}
    \item[] Question: Does the paper provide open access to the data and code, with sufficient instructions to faithfully reproduce the main experimental results, as described in supplemental material?
    \item[] Answer: \answerYes{}
    \item[] Justification: Yes. To ensure full reproducibility, our released code repository includes: the exact experimental configurations, implementations of our models and baselines, the parallel-scan implementation, scripts for generating the synthetic data, and the data preparation pipeline for language modeling.

\item {\bf Experimental setting/details}
    \item[] Question: Does the paper specify all the training and test details (e.g., data splits, hyperparameters, how they were chosen, type of optimizer) necessary to understand the results?
    \item[] Answer: \answerYes{}
    \item[] Justification: The full details of the experimental setting are provided in Appendix and codebase.

\item {\bf Experiment statistical significance}
    \item[] Question: Does the paper report error bars suitably and correctly defined or other appropriate information about the statistical significance of the experiments?
    \item[] Answer: \answerYes{}
    \item[] Justification: For all synthetic tasks, we evaluate the models using 5 different random initialization seeds. To capture the variability arising from initialization, we report the sample mean performance alongside the 1-sigma standard deviation. Furthermore, to ensure robust and fair comparisons, the "best results" reported for these synthetic tasks are selected from an extensive sweep over hidden dimensions (ranging from 100 to 6000), including hyperparameter tuning for all external baselines. For the language modeling experiments, we report the best performance achieved across a sweep of several learning rates. The aggregation methods and hyperparameter selection criteria are explicitly described in either the main text, Tables or Appendix.

\item {\bf Experiments compute resources}
    \item[] Question: For each experiment, does the paper provide sufficient information on the computer resources (type of compute workers, memory, time of execution) needed to reproduce the experiments?
    \item[] Answer: \answerYes{},
    \item[] Justification: The paper provides the amount of total compute required for synthetic tasks and language modeling experiments in Appendix
    
\item {\bf Code of ethics}
    \item[] Question: Does the research conducted in the paper conform, in every respect, with the NeurIPS Code of Ethics \url{https://neurips.cc/public/EthicsGuidelines}?
    \item[] Answer: \answerYes{}
    \item[] Justification: The authors of this paper conform to the NeurIPS Code of Ethics in every respect.

\item {\bf Broader impacts}
    \item[] Question: Does the paper discuss both potential positive societal impacts and negative societal impacts of the work performed?
    \item[] Answer: \answerNA{}
    \item[] Justification: The primary goal of this work is to advance foundational research in machine learning. As such, the proposed architectures are not currently tied to any particular downstream applications that present direct or immediate societal impacts.
    
\item {\bf Safeguards}
    \item[] Question: Does the paper describe safeguards that have been put in place for responsible release of data or models that have a high risk for misuse (e.g., pre-trained language models, image generators, or scraped datasets)?
    \item[] Answer: \answerNA{}
    \item[] Justification: The paper poses no such risks.

\item {\bf Licenses for existing assets}
    \item[] Question: Are the creators or original owners of assets (e.g., code, data, models), used in the paper, properly credited and are the license and terms of use explicitly mentioned and properly respected?
    \item[] Answer: \answerYes{}
    \item[] Justification: Yes. All the codebases, models, and data are properly credited in the text. All used assets are allowed for common use, distributed by CC-BY 4.0. Models: \cite{gu2023mamba}, \cite{siems2025deltaproduct}, \cite{yang2024parallelizing}. Code: \cite{poli2024mechanistic}, \cite{siems2025deltaproduct}, \cite{ajroldi2024plainlm}. Data: \cite{poli2024mechanistic}, \cite{merrill2024illusion}, \cite{penedo2024fineweb}.

\item {\bf New assets}
    \item[] Question: Are new assets introduced in the paper well documented and is the documentation provided alongside the assets?
    \item[] Answer: \answerYes{}
    \item[] Justification: The code is released via forks of existing codebases, accompanied by complete documentation for all introduced models and experimental configurations.

\item {\bf Crowdsourcing and research with human subjects}
    \item[] Question: For crowdsourcing experiments and research with human subjects, does the paper include the full text of instructions given to participants and screenshots, if applicable, as well as details about compensation (if any)? 
    \item[] Answer: \answerNA{}
    \item[] Justification: The paper does not involve crowdsourcing nor research with human subjects.

\item {\bf Institutional review board (IRB) approvals or equivalent for research with human subjects}
    \item[] Question: Does the paper describe potential risks incurred by study participants, whether such risks were disclosed to the subjects, and whether Institutional Review Board (IRB) approvals (or an equivalent approval/review based on the requirements of your country or institution) were obtained?
    \item[] Answer: \answerNA{}
    \item[] Justification: The paper does not involve crowdsourcing nor research with human subjects.

\item {\bf Declaration of LLM usage}
    \item[] Question: Does the paper describe the usage of LLMs if it is an important, original, or non-standard component of the core methods in this research? Note that if the LLM is used only for writing, editing, or formatting purposes and does \emph{not} impact the core methodology, scientific rigor, or originality of the research, declaration is not required.
    \item[] Answer: \answerNA{}
    \item[] Justification: LLMs were used only for writing, editing, or formatting purposes and does \emph{not} impact the core methodology, scientific rigor, or originality of the research.

\end{enumerate}

\end{document}